%% file: main.tex
\pdfoutput=1
\documentclass[11pt]{article}
\usepackage{tikz}
\input{preamble/packages}
\input{preamble/definition}

\usepackage{scalerel}

\tcbuselibrary{breakable}

\title{
\scalerel*{\includegraphics{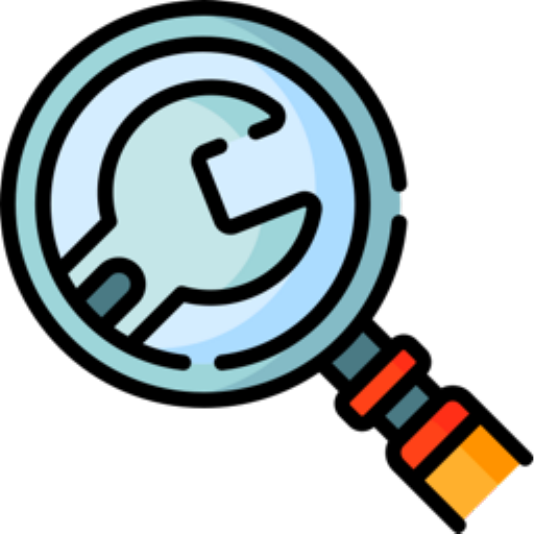}}{{\rule{2ex}{2ex}}}\hspace{-2pt}
Retrieval Models Aren't Tool-Savvy:\\Benchmarking Tool Retrieval for Large Language Models
}
\author{
Zhengliang Shi$^1$~~Yuhan Wang$^1$~~Lingyong Yan$^2$\\ \textbf{Pengjie Ren$^1$~~Shuaiqiang Wang$^2$~~Dawei Yin$^2$~~Zhaochun Ren\textsuperscript{$3$}\thanks{Corresponding author.}} \\
$^1$Shandong University, Qingdao, China~~
$^2$Baidu Inc., Beijing, China\\
$^3$Leiden University, Leiden, The Netherlands \\
[2pt]{\faGithub~\href{https://github.com/mangopy/tool-retrieval-benchmark}{\texttt{Tool-Retrieval-Benchmark}} }\\
[2pt]{\texttt{shizhl@mail.sdu.edu.cn~~z.ren@liacs.leidenuniv.nl}}
}

\begin{document}
\maketitle

\begin{abstract}

Tool learning aims to augment large language models (LLMs) with diverse tools, enabling them to act as agents for solving practical tasks.
Due to the limited context length of tool-using LLMs, adopting information retrieval (IR) models to select useful tools from large toolsets is a critical initial step.
However, the performance of IR models in tool retrieval tasks remains underexplored and unclear.
Most tool-use benchmarks simplify this step by manually pre-annotating a small set of relevant tools for each task, which is far from the real-world scenarios.
In this paper, we propose \ours, a heterogeneous tool retrieval benchmark comprising 7.6k diverse retrieval tasks, and a corpus of 43k tools, collected from existing datasets.
We benchmark six types of models on \ours.
Surprisingly, even the models with strong performance in conventional IR benchmarks, exhibit poor performance on \ours.
This low retrieval quality degrades the task pass rate of tool-use LLMs.
As a further step, we contribute a large-scale training dataset with over 200k instances, which substantially optimizes the tool retrieval ability of IR models.\footnote{Resource is available on \includegraphics[width=0.3cm]{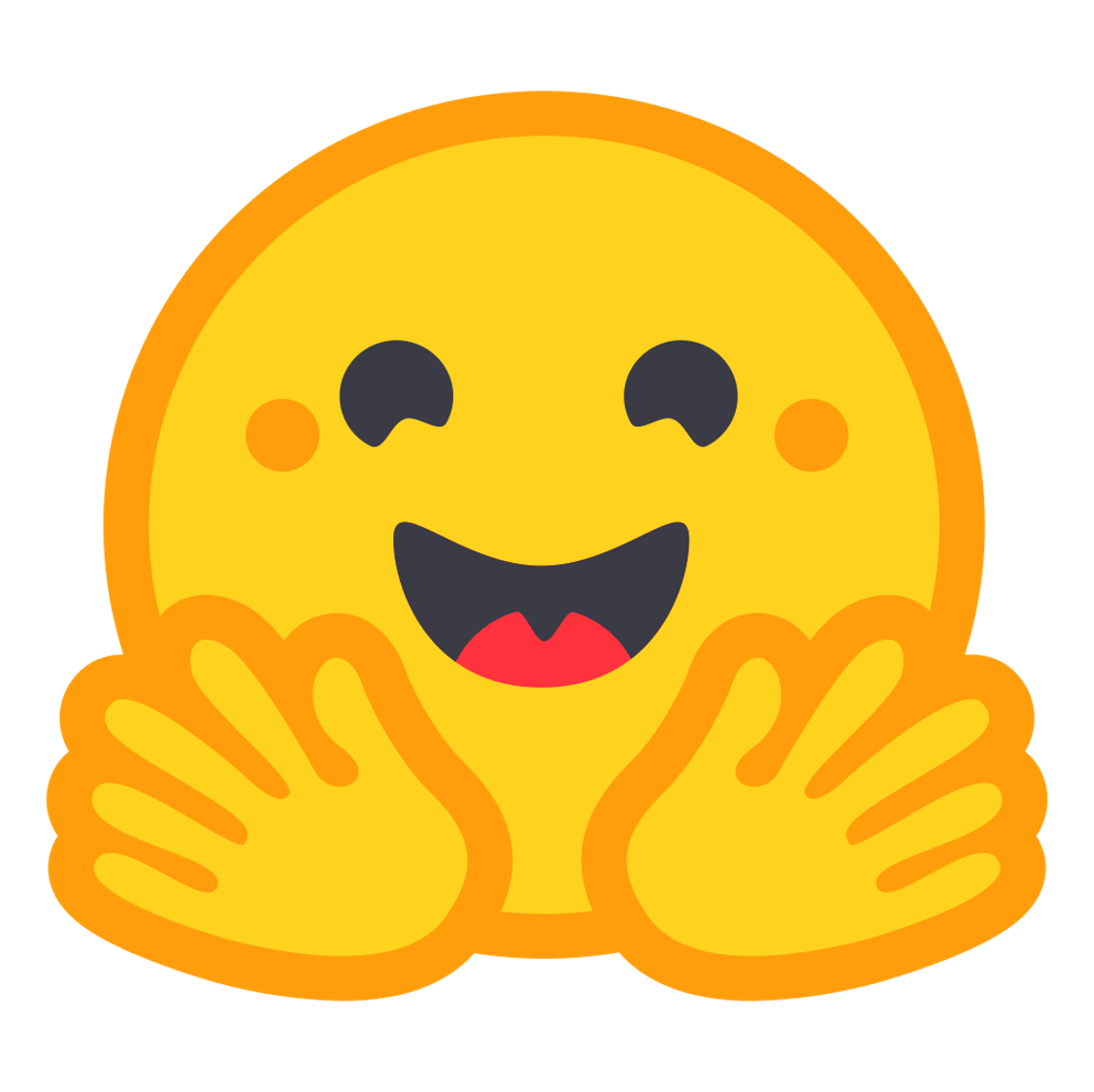} \href{https://huggingface.co/spaces/mangopy/ToolRet-demo}{Huggingface} and \faGithub~\href{https://mangopy.github.io/tool-retrieval-benchmark/}{Website}.}
\end{abstract}

\input{section/01-introduction}
\input{section/02-related-work}
\input{section/03-data-collection}
\input{section/04-dataset-analysis}
\input{section/05-experiment-setup}
\input{section/06-experiment-results}
\input{section/07-conclusion}

\section*{Limitation}
The limitations of this work include the lack of exploration in multilingual retrieval settings. Currently, our benchmark is confined to the English language and focuses exclusively on text retrieval. To address this limitation, we plan to expand our research to encompass multilingual information retrieval (IR) scenarios in the future. Additionally, another limitation lies in the insufficient investigation of prompt sensitivity. Given that large language models (LLMs) are highly sensitive to prompt wording, we aim to annotate a broader range of instructions in the future to examine how variations in prompt phrasing influence LLM performance.

Building upon \ours, we suggest the following directions for \textbf{future work}.
\begin{enumerate}[label=(\roman*)]
    \item Investigating sensitivity to instructions: Conduct a comprehensive study on how LLM performance varies with different prompt formulations and instruction styles.
    \item  In this work, we focus on the retrieval-then-calling method, where the tool retrieval step is triggered only once initially, and the top-k retrieved tools are then passed to the tool-use LLMs. In the future, we aim to \textit{benchmark} IR models in interleaved retrieval-and-calling scenarios. Specifically, the LLM can generate long-form reasoning for task planning, and the IR model can receive this reasoning output, exploring step-by-step integration between the LLM and IR models.
    \item  Enhancing IR models for improved retrieval accuracy: Further optimize IR models to achieve higher retrieval precision, leveraging these improvements to augment tool-use LLMs and, consequently, enhance end-to-end task performance. 
\end{enumerate}

\section*{Ethics Statement}
We acknowledge the importance of the ACM Code of Ethics and fully agree with it.
We ensure that this work is compatible with the provided code in terms of publicly accessible datasets and models. Risks and harms associated with large language models include the generation of harmful, offensive, or biased content. The new benchmark is composed of various previous datasets and is therefore licensed under their respective data licenses.

In this research, we prioritize reproducibility by not only utilizing state-of-the-art commercial LLMs but also experimenting extensively with open-source LLMs. Throughout the study, we have strictly followed ethical standards to maintain the integrity and validity of our work. All tools and resources used in this research were obtained from publicly available platforms, ensuring transparency and reproducibility in our experimental procedures. 
Furthermore, we have made every effort to ensure that our research does not harm individuals or groups, nor does it involve any form of deception or potential misuse of information.

\newpage
\bibliography{custom}

\input{section/appendix}


\end{document}

%% file: preamble/packages.tex
\usepackage{acl}
\usepackage{times}
\usepackage{latexsym}
\usepackage[T1]{fontenc}
\usepackage[utf8]{inputenc}
\usepackage{microtype}
\usepackage{inconsolata}
\usepackage{fontawesome}
\usepackage{times}
\usepackage{latexsym}
\usepackage{xspace}
\usepackage[T1]{fontenc}
\usepackage[utf8]{inputenc}
\usepackage{microtype}
\usepackage{inconsolata}
\usepackage{makecell}
\usepackage{amsmath}
\usepackage{amssymb}
\usepackage{amsfonts}
\usepackage{bbold}
\usepackage{multicol}
\usepackage{multirow} 
\usepackage{pifont}
\usepackage{tabularx}
\usepackage{lipsum} 
\usepackage{longtable}
\usepackage{tabularx} 
\usepackage{array} 
\usepackage{xltabular}
\usepackage{listings}
\usepackage{xcolor}
\usepackage{color}
\usepackage{colortbl}
\usepackage{minitoc}
\usepackage[inline]{enumitem}
\usepackage{booktabs,caption}
\usepackage[flushleft]{threeparttable}
\usepackage[most]{tcolorbox}
\usepackage{arydshln}

\usepackage{mathtools}
\usepackage{fixmath}
\usepackage{graphicx} %
\usepackage{subfigure}
\usepackage{xcolor}
\usepackage{siunitx}
\usepackage{geometry}
\usepackage{acronym}
\usepackage{soul}
\usepackage{adjustbox}
\usepackage[ruled,vlined]{algorithm2e}

%% file: preamble/definition.tex
\AtBeginDocument{%
  }

\definecolor{lemon}{HTML}{FDFFCC}

\definecolor{Gainsboro}{rgb}{0.86, 0.86, 0.86}

\definecolor{Gray}{gray}{0.95}
\definecolor{LightCyan}{rgb}{0.88,1,1}
\definecolor{dm-blue-500}{RGB}{0, 69, 177}
\definecolor{dm-purple-500}{RGB}{105,50,230}
\definecolor{dm-red-500}{RGB}{255,122,122}

\definecolor{backred}{RGB}{255, 190, 190}
\definecolor{backblue}{RGB}{220, 230, 250}

\newtcbox{\hlprimarytab}{on line, rounded corners, box align=base, colback=backblue, colframe=white, size=fbox, arc=3pt, before upper=\strut, top=-2pt, bottom=-4pt, left=-2pt, right=-2pt, boxrule=0pt}
\newtcbox{\hlsecondarytab}{on line, box align=base, colback=backred, colframe=white, size=fbox, arc=3pt, before upper=\strut, top=-2pt, bottom=-4pt, left=-2pt, right=-2pt, boxrule=0pt}

\newcommand{\ours}{\textsc{ToolRet}\xspace}

\newcommand{\code}[1]{{\ttfamily#1}}

\newcommand{\headernodot}[1]{\noindent\textbf{#1}}
\newcommand{\header}[1]{\headernodot{#1.}}

\acrodef{rag}{retrieval-augmented generation}
\acrodef{RAG}{Retrieval-augmented Generation}

\newcommand{\besttext}[1]{{\hlsecondarytab{#1}}}
\newcommand{\hightext}[1]{{\hlprimarytab{#1}}}

\newcommand{\high}{\cellcolor{backblue}\textbf}

%% file: section/01-introduction.tex
\section{Introduction}\label{sec:intro}

Large language models (LLMs) have demonstrated remarkable progress across various natural language processing (NLP) tasks, such as text summarization~\cite{chang2023survey}.
However, they suffer from inherent inabilities to interact with the physical world and access vast, up-to-date knowledge~\cite{qin2024tool}.
To alleviate these drawbacks, \textit{tool learning} is proposed to equip LLMs with external tools, augmenting them as agents to manipulate tools for practical task-solving~\cite{qu2025tool, wang2024tools}.

\begin{figure}[!t]
 \centering
\includegraphics[width=0.95\columnwidth]{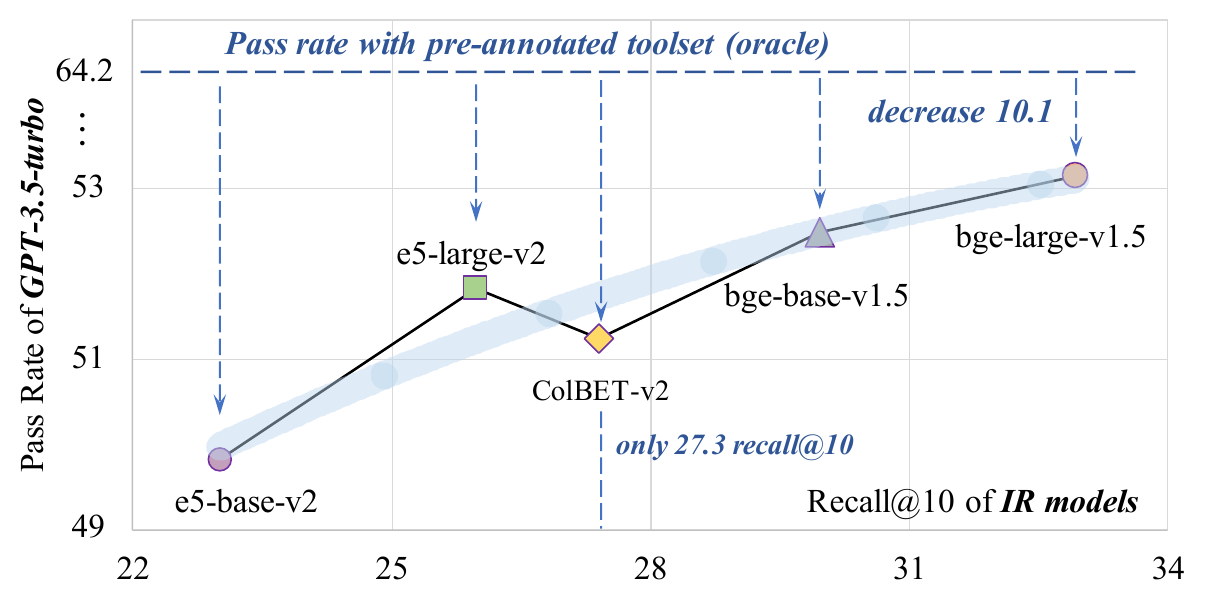}
\caption{Correlation between the tool retrieval performance (e.g., Recall@10) of IR models and the end-to-end task pass rate of tool-use agents.}\label{fig:intro}
\vspace{-3mm}
\end{figure}

In practical applications, retrieving useful tools from toolsets for LLM agents typically serves as the initial step~\cite{wang2024toolgen, xu2024enhancing, song2023restgpt}. 
This step becomes particularly critical in real-world scenarios where the candidate tools are usually large-scale and many of them are similar in functionality~\cite{qu2024colt}. 
However, most existing work~\cite{guo2024stabletoolbench, qian2023toolink} simplifies this retrieval process by manually pre-selecting a small set of 10-20 relevant tools for each evaluation task.
For example, the ToolACE~\cite{toolace} and ToolBench~\cite{qin2023toolllm} annotate about 10 tools per task.
While recent information retrieval (IR) techniques such as semantic matching~\cite{qu2024colt,xu2024enhancing}, can assist with tool retrieval, they are often trained on ad-hoc tool-use datasets, lacking comprehensive evaluation on diverse scenarios, especially for unseen tasks.
To further explore the importance of tool retrieval, we conduct a pilot experiment on  ToolBench~\cite{qin2023toolllm}.
As shown in Figure~\ref{fig:intro}, we observe that (i) the agent's performance substantially drops when replacing the officially annotated toolset with the retrieved tools; and 
(ii) even strong retrievers like colbertv2~\cite{colbertv2}, struggle to retrieve target tools effectively.
These findings highlight the necessity to (i) systematically evaluate IR models on diverse tool retrieval tasks; and (ii) analyze the impact of retrieval on the end-to-end task pass rate.

In this work, we introduce \ours, the \textit{first large-scale tool retrieval benchmark comprising 7.6k diverse retrieval tasks and a corpus of 43k tools}, which comprehensively evaluates IR models across diverse retrieval scenarios.
Specifically, we collect query-tool datasets from the following sources: (i) Tool-use agent benchmarks from published research papers in AI conferences, such as ACL and NeurIPS; (ii) Related conference resources such as AppBench in EMNLP and ToolLens in CIKM; and (iii) Other publicly available datasets from the open-source community, e.g., HuggingFace.
The collected data is carefully curated to cover a wide range of practical tool requirements, comprising diverse types of tool documentation, domains, and varying query lengths.
Then, we standardize the format of all the collected tasks, aligning them with retrieval tasks similar to the format in \textbf{MTEB}, where \textit{each retrieval task contains a query and target tools (e.g., labels)}.
To support the instructional retrieval~\cite{weller2024followir} setting of our benchmark, we also introduce a target-aware strategy to supplement each query with an instruction using the powerful LLMs (i.e., gpt-4o).

We systematically evaluate five types of IR models such as embedding models and LLM re-ranking, under various experimental settings.
Our results reveal that even the best model (i.e., \code{NV-embedd-v1}) that demonstrates strong performance in conventional IR benchmarks, achieves an nDCG@10 of only 33.83 in our benchmark.
This highlights the challenges of the tool retrieval tasks.
We identify two key factors contributing to this performance gap:
(i) Lower term overlap between queries and target tools in tool retrieval tasks, which demands higher representation abilities for IR models to accurately match query intent with the correct tools; and
(ii) Task shift from conventional information-seeking tasks (e.g., document retrieval) to tool retrieval, leading to suboptimal performance of IR models that are not explicitly optimized.

To enhance the retrieval performance and enable IR models to augment tool-use agents, we further propose the \ours-train, a large-scale training dataset containing more than 200k retrieval tasks.
We extend our data collection process from \ours to include the training set of three mainstream tool-use datasets, including ToolACE~\cite{toolace}, APIGen~\cite{liu2024apigen} and ToolBench~\cite{qin2023toolllm}.
To enable the training, we pair each retrieval task with 10 negative tools retrieved by the \code{NV-embed-v1}.
Finally, each training example contains the query, an generated instruction, the target tools, and the negative tools.
Results show that the IR models trained over \ours-train, exhibit significant improvements in the retrieval process, leading to a higher end-to-end task pass rate when integrated with tool-use LLMs.

Our contributions are summarized as follows:
(i) We introduce \ours, the \textit{first evaluation benchmark} for tool retrieval tasks.
(ii) We evaluate the tool retrieval performance of various IR models and analyze the impact of retrieval on the end-to-end task pass rate of tool-use LLMs; and
(iii) We contribute to a large-scale training dataset that enhances the performance of IR models, improving their ability to augment tool-use LLMs effectively.

%% file: section/02-related-work.tex
\section{Related work}\label{sec:related-work}

\header{Tool learning with foundation models}
Tool learning aims to equip LLMs with tools, such as web API~\cite{song2023restgpt} and python packages~\cite{codeact}, expanding their utility~\cite{qin2023toolllm}.
Existing work teaching LLMs to use tools can be broadly classified into tuning-free~\cite{chameleon} and tuning-based methods~\cite{gao2024confucius}.
The former prepends the description of candidate tools in the LLMs' context, prompting them to select and invoke tools~\cite{huang2023metatool}.
The latter enables LLMs to learn the usage of each tool through training on synthetic data~\cite{toolace, gao2024confucius}.
However, both two paradigms struggle when facing the large-scale toolset in practice~\cite{qu2024towards, liu2024apigen}.
First, real-world toolsets are typically massive, making it less possible to incorporate all tools within the limited context of LLMs.
For example, the RapidAPI platform contains more than 52k tools while the PyPI\footnote{\url{https://pypi.org/}} hosts over 600k frequently updated packages.
Second, since tools are frequently updated, it is cost-intensive to re-train the LLMs to memorize all tools~\cite{quexplore}.
Although recent studies address this challenge using semantic retrievers~\cite{qin2023toolllm,wang2024toolgen}, these solutions are typically ad-hoc and lack systematic evaluation across diverse tool retrieval scenarios.
To fill this gap, we present the \textit{first comprehensive tool retrieval benchmark} with systematic analysis.

\header{Information retrieval benchmark}
Conventional information retrieval (IR) benchmarks are typically designed for information-seeking tasks, such as Nature Question~\cite{kwiatkowski-etal-2019-natural} for question answering and MS-MARCO~\cite{bajaj2016ms} for passage re-ranking.
Recent work also explores the IR technique in various downstream tasks, such as table retrieval~\cite{chen2024tablerag,zhang2020web} and scientific retrieval~\cite{ajith2024litsearch}, which substantially augments the downstream models.
However, tool retrieval, a crucial step for tool-use agents, remains underexplored.
Compared with traditional IR tasks, retrieving useful tools is more challenging since solving a task typically requires the combination of multiple tools~\cite{qu2024towards}.
Most existing benchmarks simplify this retrieval process by manually annotating a small set of tools that fit the LLMs' context, which is far from reality with a large toolset.
In this work, we evaluate IR models on diverse tool retrieval tasks and contribute over 200k training data to facilitate future research.

%% file: section/03-data-collection.tex
\section{Benchmark construction}\label{sec:construction}


\subsection{Data collection}\label{sec:data-collection}

To build a comprehensive benchmark for tool retrieval evaluation, we collect data from the following well-known sources:
(i) \textit{Tool-use LLM benchmarks}: A wide range of benchmarks  published in leading AI conferences such as ACL and NeurIPS;
(ii) \textit{Conference Resources}: Datasets from resource tracks in IR and NLP conferences (e.g., CIKM and EMNLP); and
(iii) \textit{Other high-quality dataset}: We identify related datasets released on open-source platforms like HuggingFace and their technique reports can be found in public submissions like arXiv. We include them to enrich \ours.

Given the rapid development of benchmarks from these sources, we collect datasets released between the \textit{August 2023 to December 2024} in this version.\footnote{Our team will maintain and update the benchmark.}
We download these data from official channels based on their usage requirements and totally collect more than 30 datasets.
Since the data sources are diverse and their original formats vary substantially, we perform necessary data cleaning operations such as deduplication and text normalization to ensure consistency and quality.

We observe that most of the collected datasets are originally designed to evaluate the tool-use capability of LLMs, where the LLM is required to correctly call a sequence of target tools given an input query.
To facilitate retrieval evaluation in \ours, we align the format of all collected tasks with the well-established IR benchmark like \textit{BEIR} and \textit{MTEB}.
Specifically, each task consists of a query as input and target tools as label  (\textit{a.k.a}, ground truth), where a tool is identified by a unique identifier and paired with detailed documentation to describe its functionality.
Endpoints of the collected datasets and concrete examples of our formatted dataset are provided in Appendix~\ref{sec:app:benchmark}.


\subsection{Data sampling}\label{sec:data-sampling}
After collecting the datasets, we observe data size imbalances across different datasets.
Besides, some datasets are extremely large with substantial redundant content, making comprehensive model evaluation both inefficient and unnecessary.
Therefore, we streamline them through effective data sampling while maintaining its evaluation integrity.

\header{Task sampling}
For each collected dataset, we encode the tasks using the embedding model, i.e., \textit{NV-embedd-v1}, and apply the \textit{K-means} clustering algorithm on the text embeddings.
We set the number of clusters to the size of the corresponding toolset and randomly sample one task from each cluster.
If the toolset size exceeds the number of queries, we retain all queries.
For example, the original ToolEyes~\cite{ye2024tooleyes} dataset contains 500 queries and 95 tools; Thus, we set the cluster number as \code{min(500, 95) = 95} for clustering.

\header{Toolset sampling}
To eliminate redundancy, we manually review the documentation of each raw dataset to identify and merge identical toolsets.
For example, since the COLT~\cite{qu2024colt} toolset overlaps with the Toolbench~\cite{qin2023toolllm} , we merge their intersecting tools.
Ultimately, we merge all toolsets from the 34 datasets to form the final corpus, resulting in a total of 43k tools.
Each tool is assigned a unique identifier.

After sampling, we obtain \textit{7.6k retrieval tasks} and a corpus of \textit{43k tools}.

\subsection{Instruction construction}\label{sec:data-instruction}

Instructional information retrieval~\cite{Sun2024MAIR,weller2024followir} is an active research area, where an additional instruction is paired with the input query to guide IR models in retrieving target information.
This instruction-following capability is especially critical in tool retrieval, as IR models are often used to augment LLM agents and receive additional context from the agents beyond the input query.
To support this instructional IR scenario, we construct the instructions as part of \ours.

Considering manually writing instructions is cost-intensive and challenging to scale, we introduce a \textit{target-aware} strategy using powerful LLMs to automate this process.
Specifically, we first invite three human experts with strong NLP and IR backgrounds to manually craft 100 seed instructions.
In line with the well-defined format from~\citeauthor{asai2022task}, our instruction outlines the relevance criteria by bridging the query intent and the functionality of the target tools.
For example, for the \textit{transcribing the audio to text} task, the instruction is presented as ``\textit{retrieve tools that process audio inputs to produce accurate textual transcriptions aligned with the user requirements}''.
Next, we employ a powerful LLM, i.e., GPT-4o, as an automatic instruction generator and guide it to generate instruction for each task through in-context learning.
To enhance the diversity, we randomly sample in-context examples from the pool of both the generated and handcrafted instructions.
A detailed pseudo algorithm is provided in Appendix~\ref{sec:app:benchmark}.

After the above three processes, we obtain \ours, which consists of 7.6k tasks, each paired with an instruction, and a corpus of 43k diverse tools, providing a comprehensive testbed and supporting various evaluation settings.


%% file: section/04-dataset-analysis.tex
\section{Benchmark statistic}\label{sec:data-analysis}

\input{table/statistics}

\input{table/comparison}
\begin{figure}[t]
\centering
\includegraphics[width=1\columnwidth]{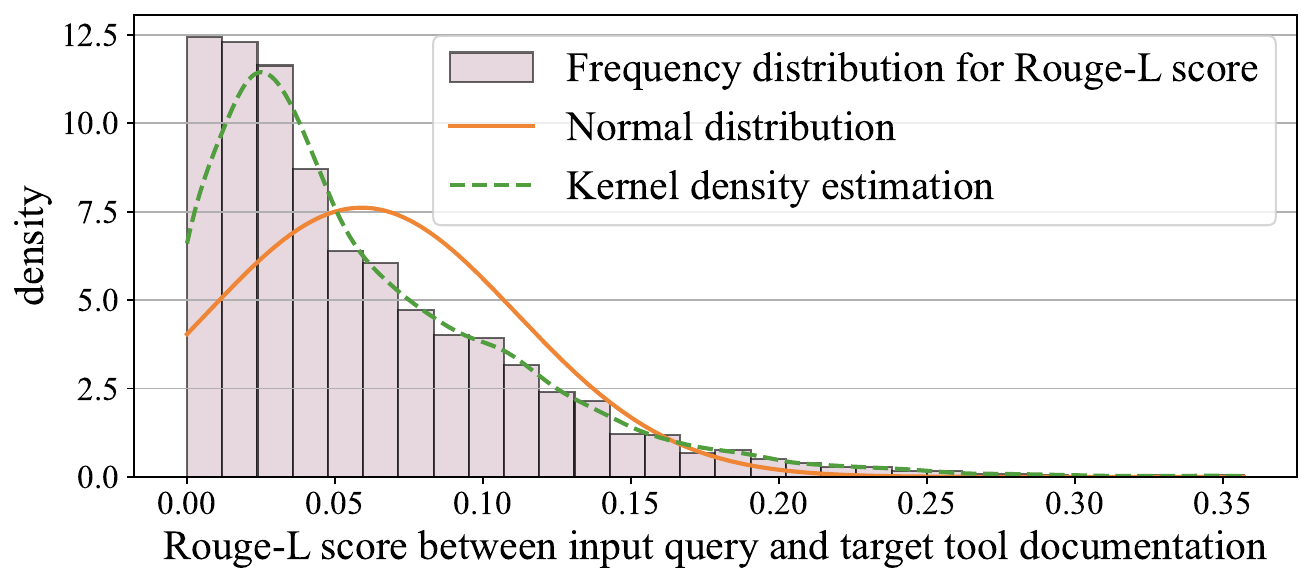} 
\caption{ROUGE-L overlap between the query (input) and the target tools (label).}\label{fig:qt}
\vspace{-2mm}
\end{figure}

Table~\ref{tab:statistics} provides the basic statics of \ours.
We observe that there are three mainstream formats of tool documentation: (i) Code, which is a function-level snippet in programming language; (ii) Web API, which elaborates the tool usage in structured JSON format following the Web OpenAPI specification; (iii) Customized application, which directly describes the tool functionality in free-form nature language.
Based on these formats, we categorize \ours into three subsets accordingly and divide the \ours into \textbf{Code Function}, \textbf{Web API}, and \textbf{Customized App} subsets.
Below, we report a more detailed analysis of \ours.

\begin{figure}[!t]
\centering
\includegraphics[width=1\columnwidth]{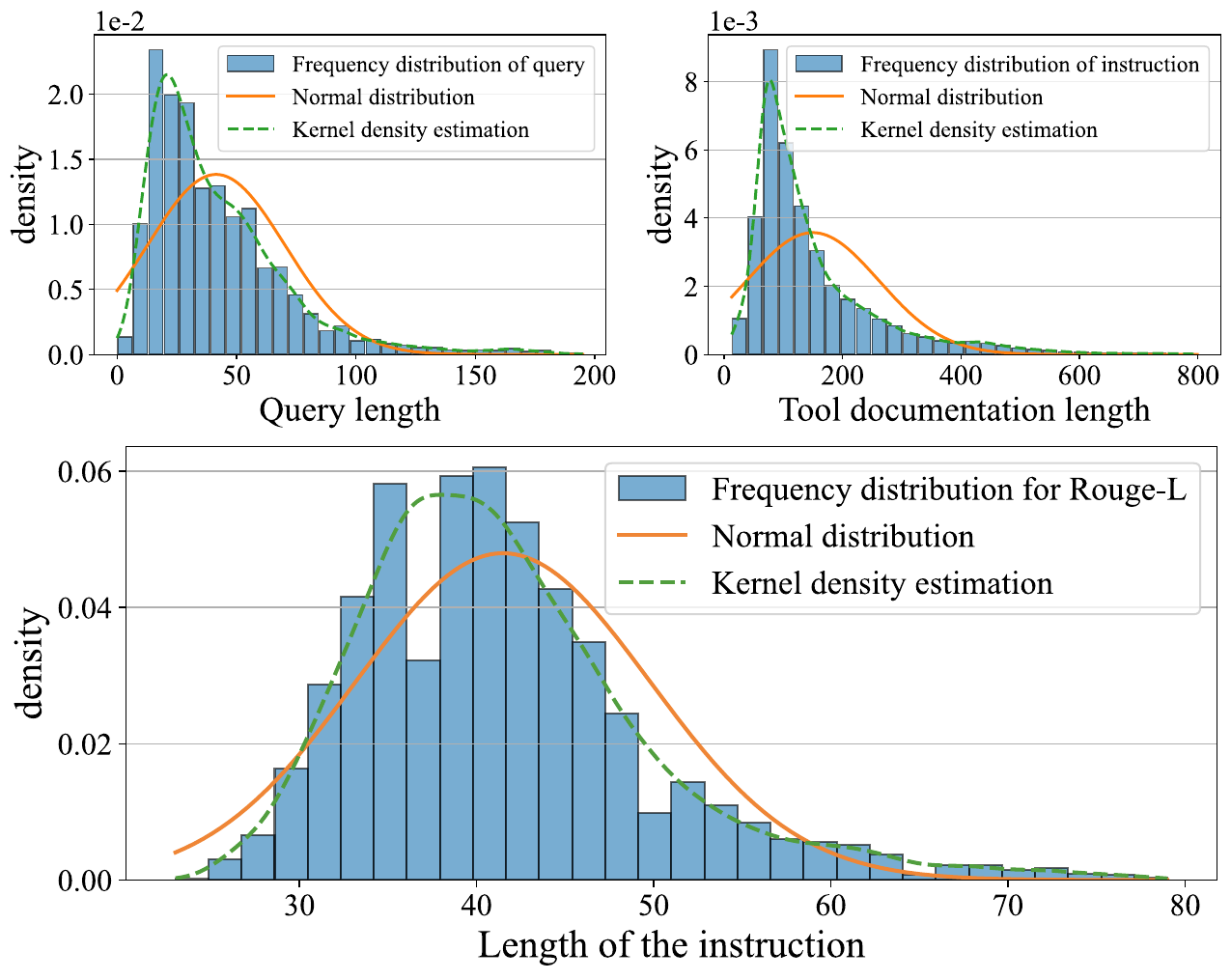} 
\caption{Length distribution of our benchmark.}\label{fig:length}
\vspace{-1mm}
\end{figure}

\subsection{Complexity}\label{sec:complexity}
In tool learning, previous studies have highlighted the necessity of combining multiple tools for task solving~\cite{shi2024chain}.
Thus, we analyze the complexity of our retrieval benchmark from two aspects.
\textbf{First}, we calculate \textit{the average number of target tools} for each retrieval task and compare it with well-known IR benchmarks such as HotpotQA~\cite{yang2018hotpotqa} and MTEB~\cite{muennighoff2022mteb}. 
As shown in Table~\ref{tab:comparison}, \ours requires models to recall more targets, posing a challenge in comprehensive retrieval.
\textbf{Second}, we compute the lexical overlap, i.e., ROUGE-L, between the input query and corresponding retrieval targets (tool documentation in \ours and passage in IR benchmarks).
We find that this overlap is substantially lower in \ours.
It indicates that, for neural IR models, \ours requires more heavily on the semantic representation rather than simple lexical matching. 
Therefore, the retrieval task in \ours is more challenging.

\subsection{Length statistics}
Figure~\ref{fig:length} illustrates the length distribution of the query, instruction, and tool documentation in \ours.\footnote{We use the tokenizer from gpt-3.5-turbo in this work.}
We find that most queries are concise, typically containing fewer than 60 tokens (about 25 words), which aligns with real-world user behavior, as users tend to input brief queries with minimal effort.
Additionally, most tool documentation is under 200 tokens,  which is similar to the chunk length in standard IR document retrieval corpus, such as Wikipedia dump~\cite{karpukhin2020dense}.

\input{table/quality}
\subsection{Quality}\label{sec:quality}
So far, we have demonstrated the complexity and quantity of our benchmark while the quality of the LLM-generated instructions remains uncertain.
To investigate this, we ask 5 human experts to label the quality based on four aspects listed in Table~\ref{tab:quality}.
Our evaluation reveals that 89.2\% of the generated instructions correctly cover the feature of the target tools and are faithfully grounded on the original queries.
For the remaining 10.8\% instructions that mismatch the query or the target tools, we ask experts to revise them.
This re-check mechanism ensures the high quality of instructions in \ours, making it a reliable evaluation benchmark.
To explain more intuitively, we list a number of seed instructions, high-quality and low-quality instructions in Table~\ref{tab:instruction-example}.
Annotation guidance is also provided in Appendix~\ref{sec:app:benchmark} to promote our transparency.

\begin{figure}[!t]
\centering
\includegraphics[width=1\columnwidth]{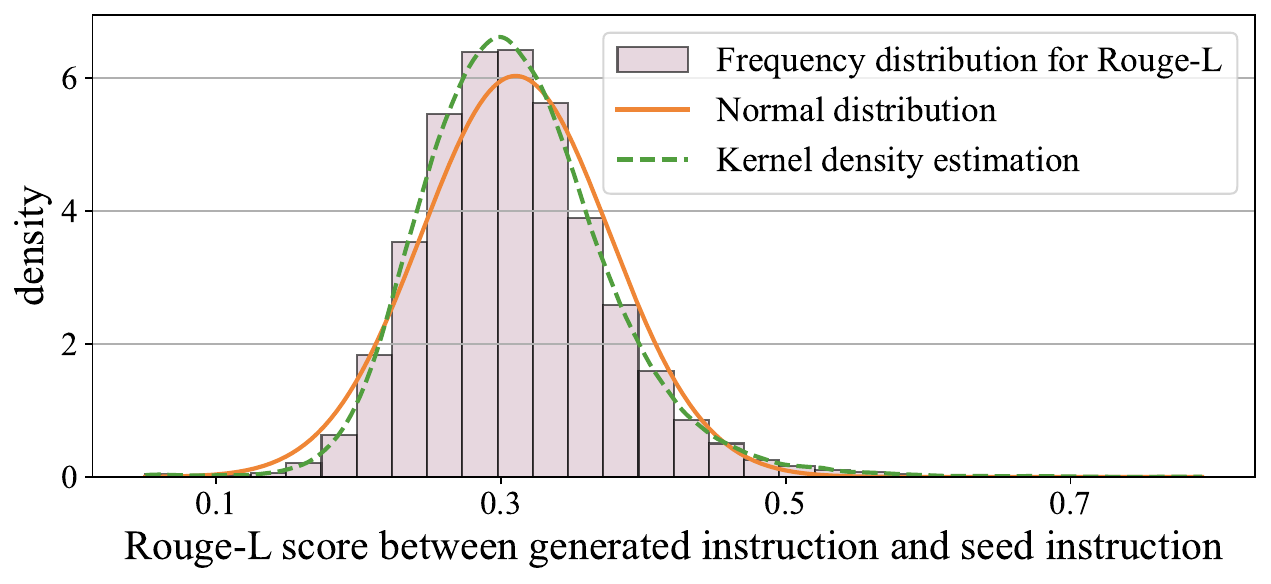} 
\caption{ROUGE-L overlap between the handcrafted seed instructions and model-generated instructions.}\label{fig:instruction}
\vspace{-3mm}
\end{figure}

\subsection{Instruction diversity}
We further analyze how the generated instructions differ from the seed instructions used to prompt the generation.
For each generated instruction, we compute its highest ROUGE-L overlap with the 100 seed instructions.
We plot the distribution of these ROUGE-L scores in Figure~\ref{fig:instruction}. 
The results indicate a decent number of new instructions are generated, which have low overlap with the seeds.


%% file: table/statistics.tex
\begin{table}[!t]
\centering
\begin{adjustbox}{width=0.9\columnwidth,center}
\begin{tabular}{@{}p{7cm} r@{}}
\toprule
\begin{tabular}[c]{@{}c@{}} \textbf{Statistic} \end{tabular}    
\\
\midrule
\# size of retrieval task &   7,615  \\
\textit{{ }} - \# of \textit{web API} retrieval task  & 4,916  \\
\textit{{ }} - \# of \textit{code function} retrieval task & 950 \\ 
\textit{{ }} - \# of \textit{customized app} retrieval task &  1,749 \\

\# size of tool & 43,215 \\
\textit{{ }} - \# of \textit{web API} &  36,978 \\
\textit{{ }} - \# of \textit{code function}  & 3,794 \\ 
\textit{{ }} - \# of \textit{customized app} & 2,443 \\
avg. query / instruction length (tokens)  & 46.87  / 43.43\\
avg. tool documentation length (token)   & 174.56 \\
\bottomrule
\end{tabular}
\end{adjustbox}
\caption{Basic statistics of our benchmark \ours.}\label{tab:statistics}
\vspace{-2mm}
\end{table}

%% file: table/comparison.tex


\begin{table}[!t]
\centering
\setlength\tabcolsep{3pt}
\begin{adjustbox}{width=0.95\columnwidth,center}
\begin{tabular}{@{}p{4cm} ccccc}
\toprule
 & Ours & NQ & MSMARCO & HotpotQA  & MTEB
\\
\midrule
\textit{\# Average number of targets for an input query.}
& \multirow{2}{*}{\hightext{\textbf{2.17}}} 
& \multirow{2}{*}{1.00} 
& \multirow{2}{*}{1.00} 
& \multirow{2}{*}{2.00} 
& \multirow{2}{*}{2.57} 
\\
\specialrule{0em}{1pt}{1pt}
\cdashline{1-6}[6pt/6pt]
\specialrule{0em}{1pt}{1pt}

\textit{\# ROUGE-L overlap between  query and targets.}
& \multirow{2}{*}{\besttext{\textbf{0.06}}}   
& \multirow{2}{*}{0.31} 
& \multirow{2}{*}{0.34} 
& \multirow{2}{*}{0.11} 
& \multirow{2}{*}{0.27} 
\\
\bottomrule
\end{tabular}
\end{adjustbox}
\caption{Comparison with conventional IR benchmarks.}
\label{tab:comparison}
\vspace{-2mm}
\end{table}

%% file: table/quality.tex
\begin{table}[!t]
\centering
\begin{adjustbox}{width=0.95\columnwidth,center}
\begin{tabular}{@{}p{7cm} c@{}}
\toprule
\begin{tabular}[c]{@{}c@{}} \textbf{Quality Review Question} \end{tabular}    
&\begin{tabular}[c]{@{}c@{}} Yes or No \% \end{tabular}    
\\
\cmidrule(r){1-1}
\cmidrule(r){2-2}

\textit{Whether the instruction is relevant to the original input query?}
& \multirow{2}{*}{ 90.1\% / 9.9\%}     
\\

\specialrule{0em}{1pt}{1pt}
\cdashline{1-2}[6pt/6pt]
\specialrule{0em}{1pt}{1pt}

\textit{Whether the instruction describes the feature of target tools}
 & \multirow{2}{*}{ 92.3\% / 8.7\%}    
\\

\specialrule{0em}{1pt}{1pt}
\cdashline{1-2}[6pt/6pt]
\specialrule{0em}{1pt}{1pt}

\textit{Whether the instruction comprehensively describe the feature of all target tools} 
&  \multirow{2}{*}{ 89.2\% / 10.8\%}   
\\

\specialrule{0em}{1pt}{1pt}
\cdashline{1-2}[6pt/6pt]
\specialrule{0em}{1pt}{1pt}

\textit{Whether the instruction contains hallucination about the target tools or input query?}
& 
 \multirow{3}{*}{ 5.9\% / 94.1\%}  
\\

\bottomrule
\end{tabular}
\end{adjustbox}
\caption{The quality review for our generated instructions, which is conducted by five human experts with 0.743 \textit{Kappa} statistics.}
\label{tab:quality}
\vspace{-2mm}
\end{table}

%% file: section/05-experiment-setup.tex
\input{table/main-without-inst}

\section{Benchmark evaluation setup}\label{sec:experiment-setup}

\subsection{Evaluation protocol}
We use three widely used IR metrics to evaluate the retrieval performance:
(i) \textit{NDCG@K} (N@K): evaluates ranking quality based on the relevance of retrieved tools;
(ii) \textit{Recall@K} (R@K): evaluates the proportion of target tools successfully retrieved within the top-K results; and
(iii) \textit{Precision@K} (P@K): evaluates the accuracy of the retrieved tools within the top-K results.
We also use \textit{Completeness@K} (C@K) from COLT~\cite{qu2024towards}, which specifically evaluates the retrieval completeness in tool retrieval tasks.
The C@K is 1 if all target tools are included in the top-k retrieved tools; otherwise, it is 0.

We mainly evaluate IR models under two settings:
(i) \textit{w/o inst.}: The model take the query alone as input; and (ii) \textit{w/ inst.}: The model takes the concatenation of the query and instruction as input to retrieve.
This allows us to analyze the impact of instructions on retrieval performance.

\subsection{Model to Evaluate}
We comprehensively evaluate the following mainstream IR models on our benchmark.

\noindent\textbf{Sparse retrieval}. These methods measure the similarity between query and tool documentation based on lexical overlap. We evaluate BM25s~\cite{L2024BM25SOO}.

\noindent\textbf{Single-task dense retrieval}. These methods use dual-encoder models trained on conventional IR datasets. We evaluate gtr~\cite{Ni2021LargeDE}, contriever~\cite{Izacard2021UnsupervisedDI}, and colbertv2.0~\cite{colbertv2}, all trained on MS-MARCO~\cite{bajaj2016ms}.
We also evaluate COLT~\cite{qu2024colt}, a recently proposed model trained on ad-hoc tool retrieval datasets.

\noindent\textbf{Multi-task embedding Models}. These methods utilize transformer encoders trained on various IR datasets. We evaluate all-MiniLM-L6-v2, gte~\cite{Li2023TowardsGT}, bge~\cite{Xiao2023CPackPR} and e5~\cite{wang2022text}, covering a wide range of sizes.

\noindent\textbf{Cross-encoder re-rankers}. These models re-rank the initially retrieved documents based on the query-passage relevance.
We evaluate: MonoT5~\cite{nogueira2020document}, mxbai-rerank-large, jina-reranker-v2-base, and bge-reranker.

\noindent\textbf{LLM agents}. These methods leverage general-purpose LLM agents for re-ranking tasks in a zero-shot setting, simulating the \textit{tool selection} process of tool-use agents. 
We evaluate the widely used LLM re-ranking framework, i.e., RankGPT~\cite{sun2023chatgpt}, with various LLMs as backbone.

Initial tools for \textit{LLM agent} and \textit{Re-ranking} baselines are retrieved by \code{Nv-embedd-v1} model.
Details of these baselines are provided in Appendix~\ref{sec:app:results}.



%% file: table/main-without-inst.tex
\begin{table*}[ht]
\centering
\begin{adjustbox}{width=2\columnwidth,center}
\setlength\tabcolsep{6pt}
\begin{tabular}{l cccc cccc cccc |cc}

\toprule
\multirow{2}{*}{\textbf{Model}} & 
\multicolumn{4}{c}{\textbf{\ours-Web}} & 
\multicolumn{4}{c}{\textbf{\ours-Code}} & 
\multicolumn{4}{c}{\textbf{\ours-Customized}} & 
\multicolumn{2}{c}{\textbf{Average}}\\
\cmidrule(lr){2-5} \cmidrule(lr){6-9} \cmidrule(lr){10-13}  \cmidrule(lr){14-15}
& N@10 & P@10 & R@10 & C@10
& N@10 & P@10 & R@10 & C@10
& N@10 & P@10 & R@10 & C@10
& N@10 & C@10\\
\midrule
BM25s  & 18.98  & 4.64  & 24.62  & 15.20  & 21.20  & 3.37  & 28.23  & 26.96  & 26.76  & 5.86  & 32.39  & 24.40  & 22.32  &  22.19  \\
\code{COLT} & 15.43   &  2.63    &   21.11   &  20.04   &   20.69	  &   5.12  &   28.07   &  18.53   &  21.63  &  4.72  &  29.19  &  22.40  &  19.25  &  20.32 \\
\code{Colbert} & 22.40  & 5.37  & 27.41  & 15.45  & 16.43  & 2.65  & 22.54  & 21.65  & 19.54  & 3.65  & 23.72  & 18.97  & 19.46  & 18.69 \\
\code{contriever-msmarco} & 21.15  & \high{5.83}  & 27.19  & 14.70  & 14.56  & 2.40  & 19.28  & 17.71  & 17.72  & 3.31  & 22.77  & 18.31  & 17.81  & 16.91 \\
\code{gtr-t5-base}   & 17.36  & 4.25  & 24.17  & 15.95  & 16.47  & 2.71  & 22.27  & 21.16  & 23.47  & 5.09  & 28.93  & 22.49  & 19.10  & 19.87 \\
\code{gtr-t5-large} & \high{22.45}  & 5.42  & \high{29.75}  & \high{18.72}  & \high{18.25}  & \high{2.89} & \high{24.12}  & \high{23.08}  & \high{26.30}  & \high{5.76}  & \high{31.86}  & \high{24.45}  & \high{22.34}  & \high{22.09} \\

\midrule
\code{all-MiniLM-L6-v2} & 11.66  & 3.07  & 16.36  & 10.15  & 14.44  & 2.50  & 19.50  & 18.11  & 22.80  & 5.21  & 29.10  & 20.25  & 16.30  & 16.17 \\
\code{e5-small-v2} & 19.89  & 5.08  & 26.46  & 16.26  & 15.48  & 2.39  & 19.26  & 18.05  & 24.60  & 5.56  & 29.67  & 20.76  & 19.99  & 18.36 \\
\code{e5-base-v2}  & 19.75  & 5.04  & 25.89  & 15.37  & 14.43  & 2.47  & 19.19  & 18.00  & 22.68  & 5.11  & 29.13  & 22.25  & 18.95  & 18.54  \\
\code{e5-large-v2}   & 18.99  & 4.90  & 25.97  & 16.27  & 17.09  & 2.68  & 21.87  & 20.70  & 26.42  & \high{6.07} & 32.19  & 23.17  & 20.83  & 20.05 \\
\code{gte-base-en-v1.5} & 23.55  & 6.28  & 32.03  & 19.15  & 17.43  & 2.87  & 23.71  & 22.48  & 21.62  & 4.76  & 29.03  & 23.17  & 20.86  & 21.60 \\
\code{gte-large-en-v1.5}& 22.41  & 5.91  & 30.14  & 18.44  & 16.66  & 2.87  & 23.64  & 22.39  & 20.62  & 5.19  & 26.75  & 17.67  & 19.90  & 19.50 \\
\code{bge-base-en-v1.5}   & 22.50  & 6.02  & 29.96  & 17.30  & 17.78  & 2.92  & 23.66  & 22.27  & \high{25.99}  & 5.71  & 32.17  & 24.26  & 22.09  & 21.27 \\
\code{bge-large-en-v1.5 } & \high{24.45}  & \high{6.66}  & \high{32.93}  & \high{19.30}  & \high{18.90}  & \high{3.12}  & \high{25.76}  & \high{24.47}  & 25.72  & 5.54  & \high{32.18}  & \high{24.79}  & \high{23.02}  & \high{22.85}  \\
\hline
\code{gte-Qwen2-1.5B-inst.}$^\spadesuit$ & 29.17 & 7.93 & 38.05 & 21.49 & 21.66 & 3.41 & 28.89 & 27.67 & 36.04 & 7.89 & 44.51 & 35.55 & 28.96  &  26.04 \\
\code{e5-mistral-7b}$^\spadesuit$  & 26.76  & 7.25  & 34.39  & 21.05  & 20.01  & 3.44  & 28.31  & 27.10  & 31.41  & 6.68  & 38.47  & 29.24  & 26.06  & 25.80  \\
\code{GritLM-7B}$^\spadesuit$ &  25.74	& 6.85  & 34.27 &	21.28	& 22.02	& 3.72	 & 30.41 & 28.87 & \high{ 42.31 }& \high{8.71} & \high{49.34} & \high{38.17} & 30.02  & 29.44 \\
\code{NV-Embed-v1}$^\spadesuit$  & \high{31.30}  & \high{8.35}  & \high{39.15}  & \high{23.05}  & \high{29.64}  & \high{4.72}  & \high{40.45}  & \high{38.88}  & 40.54  & 8.25  & 45.93  & 34.44  & \high{33.83}  & \high{32.12}  \\

\midrule
\code{mxbai-rerank-large-v1}   & 22.99  & 5.61  & 30.32  & 18.38  & 24.76  & 3.88  & 34.86  & 33.22  & 26.76  & 5.91  & 34.53  & 26.03  & 24.84  & 25.88 \\
\code{monot5-base-msmarco}  & 28.92  & 7.70  & 36.44  & 19.97  & 21.61  & 3.62  & 30.06  & 27.88  & 36.22  & 7.54  & 45.11  & 36.41  & 28.92  & 28.09 \\
\code{bge-reranker-v2-m3 } & 32.92  & 8.73  & 41.88  & 25.63  & 24.28  & 3.80  & 32.71  & 30.94  & 30.51  & 7.00  & 36.03  & 26.74  & 29.24  & 27.77 \\
\code{jina-reranker-v2-base} & 35.38  & 9.25  & 44.65  & 26.98  & 26.47  & 4.15  & 35.20  & 33.94  & 38.94  & 8.14  & 46.06  & 35.42  & 33.60  & 32.11  \\
\code{bge-reranker-v2-gemma} & \high{36.72} & \high{9.69}  & \high{45.94}  & \high{27.85}  & \high{29.89}  & \high{4.42}  & \high{38.23}  & \high{36.82} & \high{39.93}  & \high{9.06}  & \high{49.43}  & \high{37.75}  & \high{35.51}  & \high{34.14} \\

\midrule
\code{Mixtral-8x22B} & 28.21   &    \high{8.31} &  34.13  &  \high{25.42 }& 27.41 & 3.14 & 34.13 & 36.98 &  30.76 & 5.40  & 34.12& 28.65  & 28.80 &   30.35   \\
\code{gpt-3.5-turbo-0125} & 30.29    &     8.01  &    \high{36.00}    &    24.22 & 28.69  &  4.27  &  36.25  &  35.64 & 29.80 &  6.39 & 35.01   &  28.70 &  29.60 & 29.52 \\
\code{gpt-3.5-turbo-1106} &  \high{31.01}  &  7.86 &  35.82  &  23.76 &  \high{28.95} &  \high{ 4.44}  &  \high{38.16} &   \high{38.45} & \high{32.30}    &   \high{6.89} &  \high{38.31} &  \high{30.84} &  \high{30.75} &  \high{31.01} \\
\bottomrule
\end{tabular}
\end{adjustbox}
\caption{Experiment results in \textbf{\textit{w/o inst.}} setting (\S~\ref{sec:experiment-setup}), where the model takes the query as input to retrieve. We mark the baselines pre-trained on instructional datasets with $^\spadesuit$. We \hightext{highlight} the best performance in each type of model.}\label{table:main-wo-inst}
\vspace{-2mm}
\end{table*}

%% file: section/06-experiment-results.tex
\section{Experiment result}\label{sec:experiment-result}

\input{table/main-with-inst}

\begin{figure}[!t]
 \centering
\includegraphics[width=1\columnwidth]{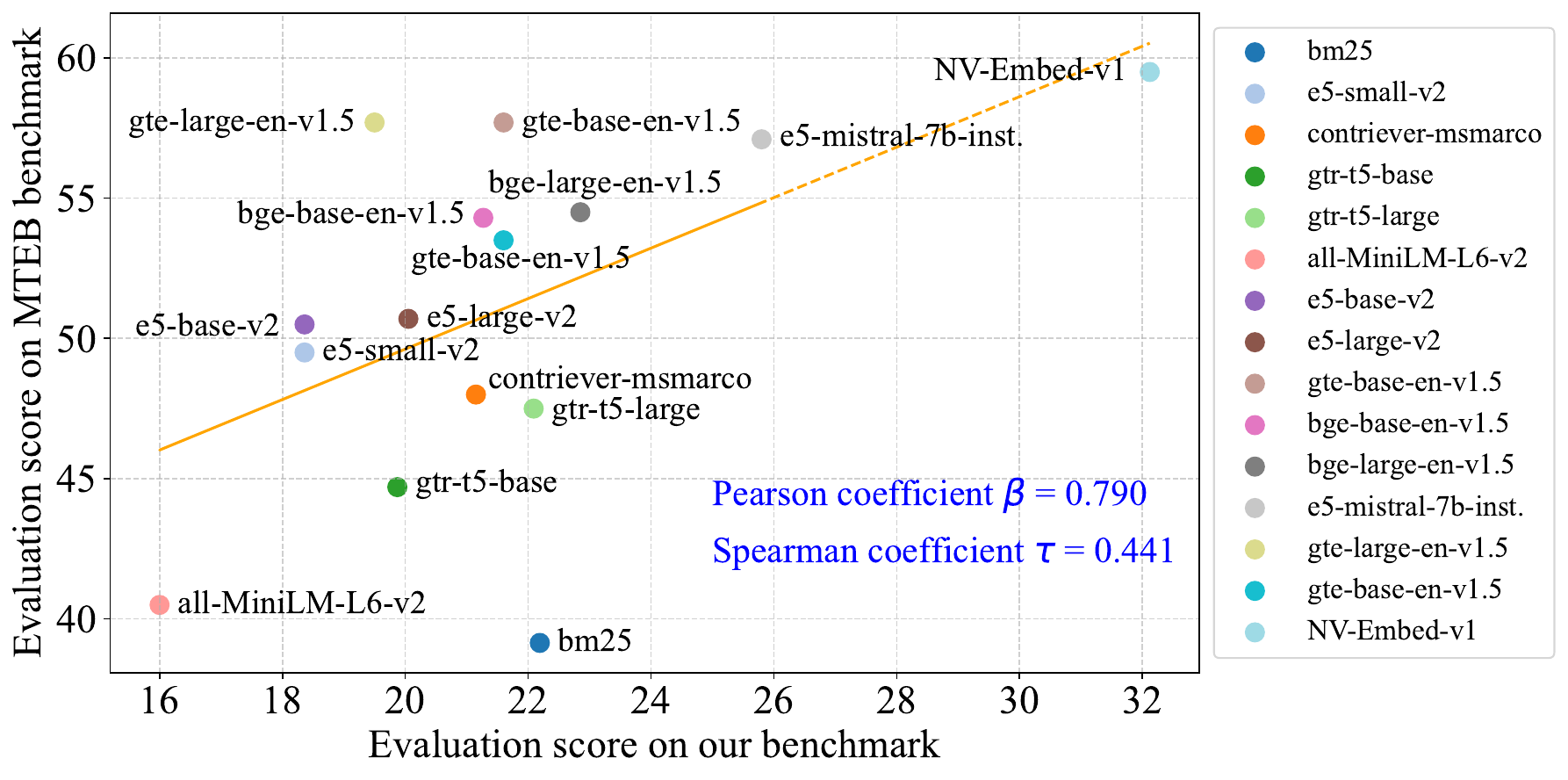} 
\caption{Correlation between the score on our benchmark and MTEB (retrieval subset).}\label{fig:correlation}
\vspace{-3mm}
\end{figure}

\subsection{Tool retrieval performance}

\header{Existing retrievers struggle}
As shown in Table~\ref{tab:main}, the tool retrieval tasks in \ours raise significant challenges for existing retriever models.
Specifically, all retrievers in our experiments achieve less than 35\% in Completeness@10 and under 52\% in recall@10.
Notably, retrieval methods that demonstrate strong performance in conventional information retrieval (IR) tasks, such as ColBERT, even underperform compared to simple lexical-based matching approaches like BM25. 
Similarly, other embedding-based models, even the NV-Embed-v1 with 7B parameter, achieve less than 45\% in completeness@10,  exhibiting limitations.

We identify two potential reasons for the above performance degradation:
(i) Tool retrieval tasks require intensive reasoning over the input query to align user intentions with candidate tools, as the lexical overlap between the query and targets is low.
(ii) There exists a domain shift between the conventional training corpora used for retrieval models and the specific tool retrieval tasks,  which current models are not explicitly optimized for.

\header{Re-ranking technique has limited improvement}
As shown in Table~\ref{tab:main}, commonly used re-ranking methods provide limited and even negative improvements for the tool retrieval task.
When using MonoT5 to re-rank the tools retrieved by NV-Embed-v1, the average NDCG@10 decreases from 33.83 to 28.92. 
A similar trend is observed with the mxbai-rerank. 
Other advanced models such as bge-ranker-v2-gemma only have 4.7\% improvement in the Completeness@10 metric.
These results further indicate the challenging nature of tool retrieval.

\vspace{-1mm}
\subsection{Substantial gains from instruction}\label{sec:gain}

Besides the evaluation results under \textit{w/o inst} setting in Table~\ref{table:main-wo-inst}, we also present the results under \textit{w/ inst} setting in Table~\ref{table:main-w-inst}.
We observe that all the IR model achieves better performance when an additional instruction is paired with the query as input.
Notably, the instruction-tuned embedding model like \code{NV-embed-v1} or \code{e5-mistral} has the most obvious improvement, which potentially benefits from its powerful instruction-following capability.
These results illustrate the advantages of the instruction and instruction tuning in tool retrieval tasks.

\begin{figure*}[t]
 \centering
\includegraphics[width=2\columnwidth]{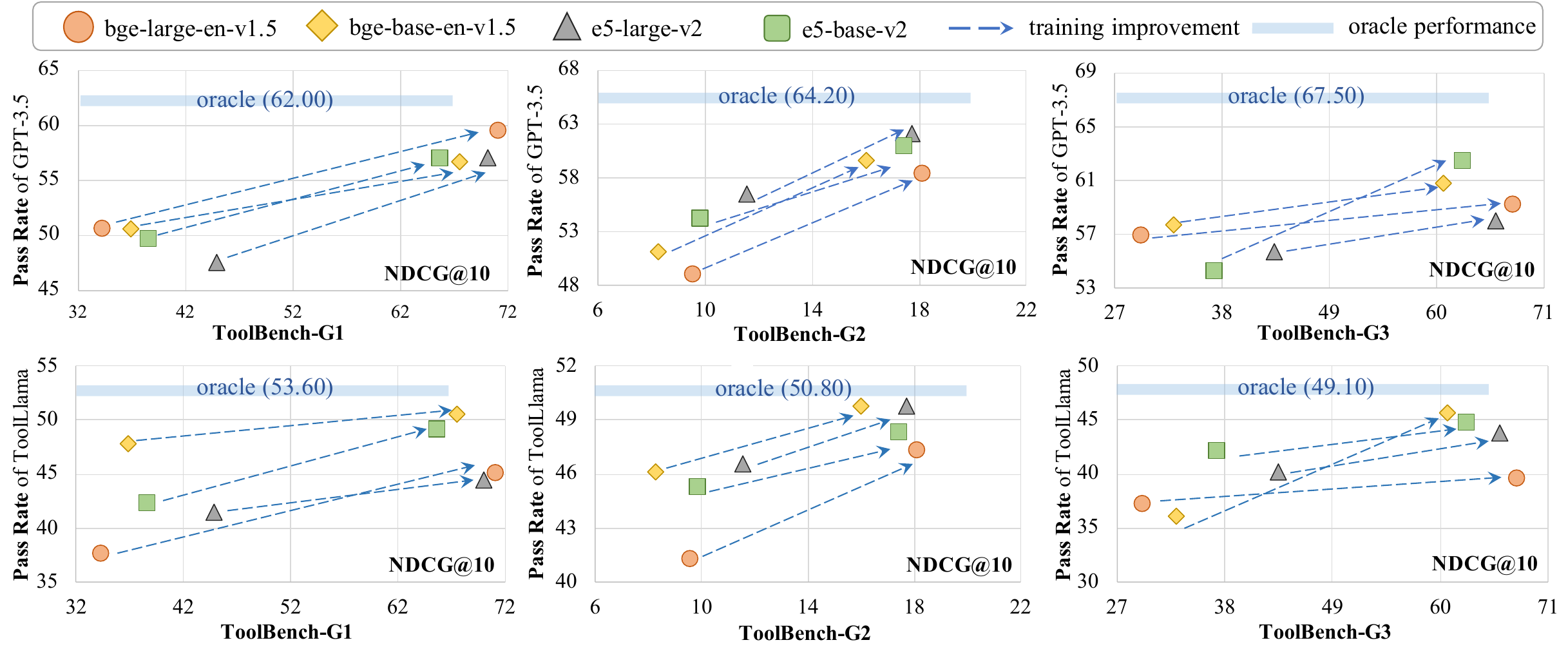}
\vspace{-2mm}
\caption{
The \textbf{horizontal axis} indicates the retrieval performance of IR models, both before and after training.
The \textbf{vertical axis} corresponds to the end-to-end pass rate of tool-use LLMs using the tools retrieved by these IR models.}\label{fig:pass}
\vspace{-2mm}
\end{figure*}

\vspace{-0.5mm}
\subsection{Compare with conventional IR tasks}

To further explore the complexity of tool retrieval tasks, we compare the models' performance on \ours and conventional IR task benchmark, i.e., MTEB, showing their relationship in Figure~\ref{fig:correlation}.
First, we can see that the two benchmarks share a similar trend (Pearson's coefficient $\beta=0.790$), but the score in \ours is lower.
This indicates that the task in \ours has a correlation with conventional IR tasks but is more challenging.
Second, we also observe that conventional IR models trained with relevance-oriented criteria such as contriever perform poorly on \ours, which indicates that \ours requires more target-aware reasoning ability.
This is also illustrated in \S~\ref{sec:complexity}.


\vspace{-1mm}
\section{Retrieval affect downstream task}\label{sec:train}

In this section, we qualitatively analyze the impact of retrieval performance on downstream tool-use agents. We conduct end-to-end evaluations on ToolBench~\cite{qin2023toolllm} dataset using the official \textit{Pass Rate} metric that evaluates whether the model successfully calls target tools to complete a task.

\subsection{Poor retrieval leads to poor tool-use agents}
For each task in ToolBench, we replace the officially pre-annotated toolset (oracle) with tools retrieved by IR models.
As shown in Figure~\ref{fig:pass}, the tool-use LLMs, when equipped with the retrieved tools, exhibit substantially lower performance compared to their oracle counterparts.
For example, in ToolBench-G1, GPT-3.5 achieves a pass rate of 50.60 using tools retrieved by \texttt{bge-large}, decreasing by 11.40.
This indicates that tool retrieval is a crucial step to build better tool-use LLMs and improve their task-solving performance.

\subsection{Towards better retrieval}
The analysis in \S~\ref{sec:gain} highlights the advantage of instruction-tuning in improving tool retrieval.
However, to the best of our knowledge, there is no large-scale instructional IR dataset for tool retrieval tasks.
We propose the \ours-train to fill this gap.

\noindent\textbf{Large-scale training data}
We extend the data collection process from \ours to include the training sets of three mainstream tool-use datasets: ToolACE~\cite{toolace}, ToolBench~\cite{qin2023toolllm} and APIGen~\cite{liu2024apigen}.
Ultimately, we collect over 200k training instances, each comprising a query $q$ and a set of target tools $\mathcal{T}$.
We also pair each query $q$ with an instruction  $\mathcal{I}$ using our target-aware strategy (See Appendix~\ref{sec:app:results}).

\noindent\textbf{Learning objective}
To train a IR model (denoted as $\theta$), we first use it to retrieved top-$K$ negative tools, denoted as $\widehat{\mathcal{T}} = \{\hat{t}_{j} \mid j\in\left[K\right], \hat{t}_{j} \notin \mathcal{T}\}$.
The model $\theta$ is then optimized by maximizing the \textit{log-likelihood of the target tools}. The loss function $\mathcal{L}$ is formulated as: 
\begin{align}
    - \frac{1}{|\mathcal{T}|} \sum_{t_{i} \in\mathcal{T}} \log  
    \frac{e^{\text{sim}(\mathcal{I}\oplus q,t_{i})}}{e^{\text{sim}(\mathcal{I}\oplus q,t_{i})} + \sum\limits_{\hat{t}_{j}\in\widehat{\mathcal{T}}} e^{\text{sim}(\mathcal{I}\oplus q,\hat{t}_{j})}}. \notag
\end{align}

The $\mathcal{I}\oplus q$ indicates concatenation of instruction and query with a special token.
During the training, we set the $K$ to 10 and the learning rate to 5e-5.

\header{Improvement from retrieval}
As shown in Figure~\ref{fig:pass}, all IR models trained on \ours-train achieve substantial improvement in NDCG@10 metric.
We further evaluate the task pass rate of two tool-use LLMs: GPT-3.5 and ToolLlama~\cite{qin2023toolllm}.
When equipped with the improved IR models, both LLMs exhibit substantial gains in pass rate, confirming the critical role of retrieval in downstream tasks.
As part of future work, we suggest adapting the IR models to better augment the tool-use LLMs, which offers a efficient plug-and-play solution compared with training LLMs.

We further conduct an ablation study by removing the instruction $\mathcal{I}$ from the loss function $\mathcal{L}$.
The results show that this variant shows improvements compared with the non-tuned counterparts, but underperforms compared with their instruction-tuned counterparts (See Appendix~\ref{sec:app:results}).
These validate the effectiveness of our instructional training data in enhancing tool retrieval performance.


\section{Discussion}

In this section, we discuss the following three open questions related to the design and reliably of our benchmark construction, as well as potential extension for future work.

\paragraph{Fine-grained functional differences of tools}

In this work, we construct the overall tool corpus by merging tools from various existing tool-use benchmarks. This heterogeneous construction strategy inevitably leads to overlapping functionality among different tools, which may raise concerns about unreliable evaluations, such as the one-to-many problem\footnote{The \textit{one-to-many} problem arises because our dataset combines multiple existing datasets. For example, for a query from dataset A, the ground truth may not be limited to the single annotation provided in A. Similar tools in dataset B might also provide valid solutions to the same query. However, in the evaluation process, only the ground truth from dataset A is used as the label, potentially leading to unreliable evaluations.}. 
However, the following reasons support the reliability of our approach:
(i) In real-world scenarios, many different but functionally similar tools exist, but typically, only the most suitable one is chosen. In this work, we focus on such real-world scenarios, where the ground truth from the original dataset is considered the most appropriate. We argue that models should have the fine-grained discrimination ability to identify the most appropriate tool;
(ii) Even if tools appear to have overlapping high-level functionality (e.g., Bing Search vs. Google Search), they often differ in important dimensions such as input parameters (e.g., support for language-specific filtering) and specific application scope (e.g., medical search vs. general news search). For example, for a query like \textit{search news articles in Chinese}, a tool like \textit{bing\_search\_with\_lang\_param} is a more precise match than a generic search API without language constraints.

In line with prior work~\cite{gao2024confucius,qin2023toolllm,quexplore}, our benchmark encourages models to retrieve semantically and functionally appropriate tools, not merely those with similar surface forms.

\paragraph{Reason for our format-based categorization}

Unlike existing tool retrieval benchmarks~\cite{qu2024colt}, we introduce a novel task categorization by dividing the overall \ours dataset into three subsets: \textit{Web APIs}, \textit{Code Functions}, and \textit{Customized Applications}. This division is motivated by two reasons:
(i) It naturally reflects the structural features of tools collected from over 30 diverse datasets. This categorization aligns with conventions established in prior research on tool-augmented LLMs, where these three formats are commonly used as the primary categories for executable tools.
(ii) This format-based taxonomy provides a clear, interpretable, and effective framework for analyzing model behavior across different tool representations.

\paragraph{Beyond format-based category}
In addition to the format-based categorization presented in this work, we believe that other dimensions of categorization can also offer valuable insights into retrieval behavior.
Therefore, we propose extending the category structure in future work to include three additional dimensions for more comprehensive and customized evaluations:
\begin{enumerate*}[label=(\roman*)]
    \item Query Length (Input Complexity): Tasks are grouped based on the query token length, with categories such as 0–25, 25–50, 50–75, and 75+ tokens.
    \item Number of Target Tools (Multi-Label Complexity): Queries are grouped by the number of relevant ground-truth tools, categorized as 1, 2–3, or $\geq$ 4 tools.
    \item Lexical Overlap (ROUGE-L) Between Query and Tool Documentation: Tasks are grouped by the degree of lexical overlap, categorized as low (<0.05), medium (0.05–0.2), and high ($\geq$ 0.2). A lower overlap indicates a higher need for deep semantic matching between the query and the correct tools.
\end{enumerate*}

%% file: table/main-with-inst.tex
\begin{table*}[ht]
\centering
\setlength\tabcolsep{6pt}
\begin{adjustbox}{width=2.0\columnwidth,center}
\begin{tabular}{l cccc cccc cccc |cc}
\toprule
\multirow{2}{*}{\textbf{Model}} & 
\multicolumn{4}{c}{\textbf{\ours-Web}} & 
\multicolumn{4}{c}{\textbf{\ours-Code}} & 
\multicolumn{4}{c}{\textbf{\ours-Customized}} & 
\multicolumn{2}{c}{\textbf{Average}} \\
\cmidrule(l){2-15}
& N@10 & P@10 & R@10 & C@10
& N@10 & P@10 & R@10 & C@10
& N@10 & P@10 & R@10 & C@10
& N@10 & C@10 \\
\midrule
\code{BM25s}  & 26.33  & 6.10  & 34.22  & 22.79  & 41.90  & 6.20  & 56.49  & 55.39  & 41.16  & 8.39  & 48.60  & 38.90  & 36.46  & 39.03 \\
\code{COLT} &28.91 &4.61  &40.64 &38.83	& 20.06& 4.71 &27.78 &18.84 & 31.29 &6.05  &42.19  &34.01 &26.75& 30.56  \\
\code{Colbert} & 16.67  & 3.12  & 21.14  & 14.94  & 30.35  & 4.37  & 41.38  & 40.28  & 24.35  & 4.56  & 30.97  & 24.87  & 23.79  & 26.70 \\
\code{contriever-msmarco}  & 23.48  & \high{5.29}  & 30.21  & 19.69  & 31.61  & 4.84  & 43.01  & 41.74  & 21.93  & 3.85  & 27.28  & 23.04  & 25.67  & 28.16 \\
\code{gtr-t5-base}   & 20.38  & 4.49  & 27.53  & 19.24  & 33.59  & 4.90  & 43.18  & 41.88  & 41.84  & 7.66  & 48.35  & 39.28  & 31.94  & 33.46 \\
\code{gtr-t5-large}  & \high{24.37}  & 5.27  & \high{31.64}  & \high{21.26}  & \high{36.76}  & \high{5.33}  & \high{47.42}  & \high{45.92}  & \high{42.04}  & \high{8.48}  & \high{50.84}  & \high{40.00}  & \high{34.39}  & \high{35.73} \\

\midrule
\code{all-MiniLM-L6-v2} & 12.77  & 3.26  & 19.38  & 13.33  & 31.59  & 5.06  & 43.86  & 42.25  & 32.24  & 7.14  & 43.55  & 32.34  & 25.53  & 29.31 \\
\code{e5-small-v2} & 26.42  & 6.20  & 34.44  & 21.39  & 32.36  & 4.84  & 42.38  & 41.11  & 34.62  & 6.90  & 42.29  & 32.58  & 31.14  & 31.69 \\
\code{e5-base-v2}  & 24.71  & 5.78  & 33.45  & 21.94  & 31.40  & 5.01  & 42.83  & 41.38  & 38.06  & 7.54  & 46.84  & 36.43  & 31.39  & 33.25 \\
\code{e5-large-v2}  & 23.62  & 5.52  & 32.19  & 21.80  & 34.27  & 5.05  & 44.42  & 43.19  & 43.32  & 8.51  & 52.30  & 41.42  & 33.73  & 35.47 \\
\code{gte-base-en-v1.5}  & \high{30.75}  & 7.00  & \high{39.44}  & \high{25.88}  & \high{41.68}  & \high{6.20}  & \high{53.96}  & \high{51.64}  & 37.95  & 6.96  & 46.57  & 38.10  & 36.79  & 38.54 \\
\code{gte-large-en-v1.5} & 28.06  & 6.55  & 36.32  & 22.57  & 35.77  & 5.75  & 49.56  & 47.71  & 37.27  & 7.88  & 47.98  & 35.84  & 33.70  & 35.37 \\
\code{bge-base-en-v1.5}  & 25.95  & 6.16  & 35.12  & 23.40  & 35.15  & 5.22  & 45.74  & 44.32  & 43.20  & \high{8.82}  & \high{53.54}  & \high{42.29}  & 34.77  & 36.67 \\
\code{bge-large-en-v1.5 } & 30.03  & \high{7.01}  & 39.28  & 25.63  & 41.53  & 6.00  & 52.76  & 51.18  & \high{43.90} & 8.31  & 51.79  & 42.24  & \high{38.49}  & \high{39.68} \\

\code{e5-mistral-7b}$^\spadesuit$  & 31.07  & 7.65  & 41.30  & 27.04  & 44.97  & 6.66  & 58.95  & 56.79  & 40.88  & 7.91  & 49.35  & 38.35  & 38.97  & 40.73 \\
\code{NV-Embed-v1}$^\spadesuit$ & 31.51  & 7.74  & 40.52  & 26.74  & \high{47.92}  & 7.10  & \high{62.07}  & \high{59.60 } & 48.70  & 10.07  & 57.69  & 43.88  & 42.71  & 43.41 \\
\code{gte-Qwen2-1.5B-inst.}$^\spadesuit$ & \high{37.53} & 9.31 & \high{48.31} & \high{30.95}	 & 47.38	& \high{7.29} & 61.12 & 59.55	 & \high{52.98} & \high{10.63}	 & \high{59.47}	 & \high{45.68} & \high{45.96} & \high{45.39} \\
\code{GritLM-7B}$^\spadesuit$ &  36.58	& \high{9.34} & 46.01	  &	27.65	& 41.26 & 6.17 	 & 53.81 & 52.07& 45.55 & 9.74 & 54.01 &41.40 & 41.13 & 40.37	\\

\midrule
\code{mxbai-rerank-large-v1}  & 17.53  & 4.05  & 25.82  & 17.95  & 33.86  & 5.05  & 47.84  & 46.47  & 26.83  & 6.71  & 37.61  & 28.60  & 26.08  & 31.01 \\
\code{monoT5-base-msmarco} & 23.33  & 5.88  & 30.70  & 18.13  & 31.39  & 5.27  & 45.18  & 42.51  & 37.77  & 6.76  & 46.63  & 39.70  & 30.83  & 33.45 \\
\code{bge-reranker-v2-m3 } & 34.83  & 8.54  & 45.23  & 31.73  & 50.86  & 7.64  & 67.26  & 64.78  & 42.35  & 9.52  & 53.75  & 39.90  & 42.68  & 45.47 \\
\code{jina-reranker-v2-base} & \high{42.35}  & \high{10.11}  & \high{51.21}  & \high{34.23}  &  53.21 & 7.66  & 66.03  & 63.94  & 45.94  & 10.36  & 57.96  & 45.41  & 47.17  & 47.86 \\
\code{bge-reranker-v2-gemma} & 34.73  & 8.09  & 45.08  & 32.29  & \high{55.85}  & \high{8.22}  & \high{70.53}  & \high{68.76}  & \high{51.97}  & \high{11.04}  & \high{61.20 } & \high{45.65}  & \high{47.52}  & \high{48.90} \\

 \midrule
\code{Mixtral-8x22B}   & 35.31 & 7.56 & 38.63 & 34.60 & 33.27 & 5.77 & 39.60 & 38.53 & 34.40 & 6.44 & 39.72 & 38.20 &34.33 &  37.11 \\
\code{gpt-3.5-turbo-0125} &  37.22    &    8.97   &     40.82    &    35.22 & 35.42  &  6.22  &    41.16  &  42.64 & 37.29  &  \high{8.24}   & 41.34   &  \high{ 39.70} &   29.60 & 29.52 \\
\code{gpt-3.5-turbo-1106} & \high{38.31}    & \high{9.02}   & \high{41.29}   & \high{35.76 }&   \high{38.69}  & \high{ 7.27}    & \high{42.57}  &   \high{42.81} &  \high{39.30}  &  7.89  &  \high{43.31}   & 37.31 &  \high{38.77} &  \high{ 38.63} \\
\bottomrule
\end{tabular}
\end{adjustbox}
\caption{Experiment results in \textbf{\textit{w/ inst.}} setting (\S~\ref{sec:experiment-setup}), where the model takes the query and instruction as input to retrieval. $^\spadesuit$ indicates the model is pre-trained on instructional datasets. We \hightext{highlight} the best performance.}\label{table:main-w-inst}
\vspace{-2mm}
\end{table*}

%% file: section/07-conclusion.tex
\vspace{-1mm}
\section{Conclusion}\label{sec:conclusion}

In this work, we introduce \ours, the first diverse tool retrieval benchmark comprising 7.6k queries, each paired with an instruction, and a corpus of 43k tools. 
\ours is a heterogeneous benchmark, constructed by aggregating existing tool-use datasets and aligning them into a unified format, similar to conventional IR benchmarks such as MTEB.
We evaluate state-of-the-art IR models on \ours and uncover a surprising finding: even models with strong performance on conventional IR benchmarks struggle in tool retrieval. This low retrieval quality significantly degrades the end-to-end task pass rate of tool-use LLMs.
Inspired by this, we further propose \ours-train, a large-scale training set containing over 200k retrieval tasks. Results show that IR models trained on \ours-train exhibit substantial improvement and also enhance the pass rate of tool-use LLMs by 10\%-20\%.
In the future, we plan to extend the \ours into multimodal scenarios.

\section*{Acknowledgements}
This work was supported by the Natural Science Foundation of China (Grant No. 62472261), the Shandong Province Key Research and Development Program (2024CXGC010108), and the Shandong Province Technology Innovation Guidance Program (YDZX2024088).

\clearpage
\newpage

%% file: section/appendix.tex
\appendix
\newpage
\onecolumn
\section{Data Card}
Following previous work~\cite{bender2018data, gebru2021datasheets, zhuo2024bigcodebench}, we provide the datacard for \ours, where we tend to summarize and centralize all information that might be relevant for the benchmark analysis.
\begin{enumerate}[label=(\roman*)]
    \item \textit{The purpose of this benchmark}: This benchmark is proposed to comprehensively evaluate the information retrieval (IR) models on tool retrieval tasks. On top of \ours, we find that existing IR models, despite achieving strong performance in conventional IR benchmarks such as MTEB and BEIR, still suffer from substantial challenges in tool retrieval tasks. The poor retrieval quality further degrades the end-to-end task pass rate of tool-use LLMs. Thus, we believe that the \ours reveals the importance of tool retrieval in building better tool-use LLMs, and can be used as a comprehensive and fair benchmark in facilitating the development of tool retrieval models.
    \item \textit{How will the dataset be distributed (e.g., Tarball on Website or Github)?} The proposed benchmark \ours will be released to the public, and hosted on GitHub and Hugging Face. The \ours will be managed and maintained by our research team.
    \item \textit{Will the dataset be updated (e.g., to correct labeling errors, add new instances, delete instances)?} Yes. If we include more tasks or find any errors, we will correct the dataset hosted on Hugging Face and GitHub and update the results in the leaderboard accordingly. It will be updated on our website.
    \item \textit{Will the training dataset \ours-train will be released publicly.}
    Yes, the proposed training dataset \ours-train will be released to the public, and hosted on GitHub and Hugging Face. 
\end{enumerate}

\section{Details of Benchmark}\label{sec:app:benchmark}

\subsection{Dataset collections}
\ours is a heterogeneous benchmark that integrates a wide range of well-established tool-use datasets and aligns them into a unified format, similar to standard information retrieval (IR) benchmarks such as \textbf{BEIR} and \textbf{MTEB}, to facilitate tool retrieval evaluation. In tool learning, we observe that previous work primarily focuses on three mainstream types of tools:

\begin{enumerate}[label=(\roman*)]
    \item \textit{Web APIs}: These tools are encapsulated in the OpenAPI format (standard JSON documentation) and can be directly invoked via HTTP requests. Web APIs are typically used to access, manipulate (e.g., add, delete, edit, or query), or retrieve private data or information from specialized databases, covering a wide range of domains such as movies, music, and sports.
    \item \textit{Code Functions}: These tools are represented by source code containing function signatures and implementation details. Code functions primarily focus on low-level computations or atomic operations, such as tensor calculations, calling Hugging Face models, or utilizing PyTorch libraries.
    \item \textit{Customized Apps}: These tools are paired with free-form natural language descriptions. They are typically user-oriented or personalized, enabling tasks such as sending emails or other custom applications.
\end{enumerate}
These tool types differ in functionality and documentation format, reflecting diverse scenarios for tool-use LLMs. For IR models, retrieving different types of tools may present varying levels of difficulty. Therefore, we categorize the collected datasets into these three types based on their paired toolset formats, resulting in three subsets of \ours: \ours-web, \ours-code, and \ours-customized. During evaluation, we report the performance of IR models on each subset to provide a fine-grained analysis. Below, we list the datasets included in each subset and provide detailed descriptions.

\subsection{\ours-Web}
The \ours-Web subset is constructed by integrating the following datasets, which contain tools in the form of Web APIs:

\begin{itemize}
    \item \textbf{AutoTools-Food}~\cite{shi2024chain}: Contains APIs related to food recipes, where LLMs must retrieve specific food-related tools to answer user queries.
    \item \textbf{RestGPT-TMDB}~\cite{song2023restgpt} and \textbf{AutoTools-Movie}~\cite{shi2024chain}: Includes web APIs from the \textbf{TMDB} platform, a movie database. Evaluation tasks require LLMs to retrieve tools to find relevant information about movies or celebrities and extract key evidence to answer given queries.
    \item \textbf{AutoTools-Weather}~\cite{shi2024chain}: Features web APIs from a weather database. LLMs must invoke these APIs and gather responses to answer weather-related queries.
    \item \textbf{RestGPT-Spotify} and \textbf{AutoTools-Music}~\cite{shi2024chain}: Contains web APIs from a music platform. Evaluation tasks require LLMs to retrieve tools for searching songs or albums based on user queries.
    \item \textbf{ToolBench}~\cite{qin2023toolllm}: Comprises over \textbf{16,000} web APIs crawled from \textbf{RapidAPI}. Queries are generated by LLMs, with ground truth tools labeled for each query.
    \item \textbf{ToolLens}~\cite{qu2024colt}: A subset of \textbf{ToolBench}, where queries are annotated to evaluate tool functionality.
    \item \textbf{APIbank}~\cite{li2023api}: Contains web APIs for daily personalized applications, such as alarm booking and database login.
    \item \textbf{MetaTool (ToolE)}~\cite{huang2023metatool}: Designed to evaluate whether LLMs are aware of tool usage and can correctly select tools.
    \item \textbf{Mnms}~\cite{ma2024m}: Evaluates LLM-based agents' tool-use abilities for \textbf{multi-step, multi-modal} tasks involving tools that process visual information. Since \ours focuses on text-based IR models, images are represented using their URLs.
    \item \textbf{Reverse-Chain}~\cite{zhang2023reverse}: Contains diverse \textbf{multi-step tasks} requiring LLMs to invoke relevant tools sequentially.
    \item \textbf{ToolEyes}~\cite{ye2024tooleyes}: Includes tools across various domains, such as \textbf{advice, entertainment, and art}, providing a broad evaluation of tool-use LLMs in practical scenarios.
    \item \textbf{UltraTool}~\cite{huang2024planning}: A benchmark designed to improve and evaluate LLMs' tool utilization abilities in \textbf{real-world scenarios}, focusing on the entire process of planning, creating, and applying tools in complex tasks.
    \item \textbf{T-Eval}~\cite{chen2023t}: A fine-grained benchmark assessing tool-use LLMs across multiple evaluation aspects, including \textbf{instruction following, planning, reasoning, retrieval, understanding, and review}.
\end{itemize}
\input{table/dataset}

\subsection{\ours-Code}
The \ours-code subset is constructed by integrating the following datasets, which contain tools in the form of code functions:

\begin{itemize}
    \item \textbf{Gorilla-PyTorch}~\cite{patil2023gorilla}: Contains various \textbf{PyTorch functions} (code snippets) as tools, evaluating LLMs' ability to correctly combine PyTorch functions for solving deep learning tasks. The functions in this dataset are collected from the \textbf{Python Torch} package.
    \item \textbf{Gorilla-Tensor}~\cite{patil2023gorilla}: Includes \textbf{TensorFlow functions} as tools, collected from \textbf{TensorFlow Hub}, to assess LLMs' tool selection capabilities in deep learning scenarios.
    \item \textbf{Gorilla-HuggingFace}~\cite{patil2023gorilla}: Treats specific \textbf{downstream models} from the \textbf{Hugging Face} platform as tools. This dataset evaluates LLMs' performance in correctly calling Hugging Face models based on user queries.
    \item \textbf{CRAFT-TabMWP}~\cite{yuan2023craft}: Evaluates LLMs' ability to use \textbf{functions for table processing}. The functions in this dataset are first generated by \textbf{GPT-4} and subsequently verified.
    \item \textbf{CRAFT-VQA}~\cite{yuan2023craft}: Provides evaluation cases for \textbf{visual question answering (VQA)}, where LLMs must call image processing functions such as \textbf{image capture} and \textbf{object detection}.
    \item \textbf{CRAFT-Math-Algebra}~\cite{yuan2023craft}: Assesses LLMs' ability to invoke \textbf{algebra functions} for solving complex mathematical problems.
\end{itemize}

\subsection{\ours-Customized}
Besides Web APIs and code functions, we also collect datasets that contain customized apps.  
Unlike Web APIs and code functions, customized apps are described using free-form natural language documentation rather than structured formats.
Specifically, we include the following datasets:  
\textbf{ToolACE}~\cite{toolace},  
\textbf{GPT4Tools}~\cite{yang2024gpt4tools},  
\textbf{TaskBench}~\cite{shen2023taskbench},  
\textbf{ToolAlpaca},  
\textbf{ToolBench-sam}~\cite{xu2023tool},  
\textbf{ToolEmu}~\cite{ruan2023identifying}, and  
\textbf{TooLink}~\cite{qian2023toolink}.

\input{table/algorithm}

\subsection{Task format}
The final benchmark, \ours, integrates the above datasets and reformats all test cases into a unified format, similar to conventional IR benchmarks such as \textbf{BEIR} and \textbf{MTEB}, to evaluate IR models in tool retrieval tasks.
Each reformatted task consists of: an input query, an instruction, and the corresponding target tools (e.g., labels).
Each tool is assigned a unique identifier and is paired with detailed documentation describing its functionality.
Below, we present a concrete example from \ours.
\input{table/case}

\subsection{Details of instruction construction}
In \ours, each task is paired with an instruction using a target-aware strategy, where GPT-4o acts as an automatic expert through in-context learning.
Specifically, we follow these steps.
We first invite three human experts with strong backgrounds in NLP and IR to manually craft seed instructions.
These expert-crafted instructions form an initial example pool.
For each task, we randomly sample a set of instructions from this pool as in-context learning examples.
Using these examples, GPT-4o generates a new instruction tailored to the given task.
The newly generated instruction is then appended back to the instruction pool to enhance instruction diversity.
The detailed pseudo algorithm is provided in Alg.~\ref{algo:instruction}.

Below, we provide a concrete example of the GPT-4o prompt used in our instruction construction process.
The example of seed instructions and the generated instructions is provided in Table~\ref{tab:instruction-example}.
\begin{lstlisting}[basicstyle=\small\ttfamily, breaklines=true, breakindent=0em, commentstyle=\color{red!50!green!50!blue!50}, frame=shadowbox, rulesepcolor=\color{red!20!green!20!blue!20},numbers=none,literate={`}{\textasciigrave}1]
# The prompt for GPT-4o.

Given a query, you need to design an instruction about 20 words that clearly indicates this is a task to retrieve tools capable of solving the query based on its content. The instruction should emphasize the task requirements and target outcomes of the query while incorporating the functional characteristics of the tools to help the system accurately match the appropriate tools.

Below, I have provided the target tools (i.e., the labels for the query). Please analyze the key aspects of the query and the tool descriptions. Your instruction should implicitly highlight the task requirements and the characteristics of the target tools relevant to the query.

Here is an output template that your should follow. Please note that the instruction should be concise.

Query: I would like to generate a video presenting a text-based discussion on the topic of 'The Benefits of Exercise'
Labels: [1] {'id': 'taskbench_data_huggingface_tool_5', 'doc': {'input-type': ['text'], 'output-type': ['text'], 'name': 'Text Generation', 'description': 'Generating text is the task of producing new text. These models can, for example, fill in incomplete text or paraphrase.'}}
Instruction: Given a `text-to-video` task, retrieve tools that process text inputs to generate coherent textual outputs aligned with the query's topic and requirements.

Query: I have an audio file 'example.wav' which is difficult to understand. I would like you to help me transcribe the audio to text
Labels: [1] {'id': 'taskbench_data_huggingface_tool_19', 'doc': {'input-type': ['audio'], 'output-type': ['text'], 'name': 'Automatic Speech Recognition', 'description': 'Automatic Speech Recognition (ASR), also known as Speech to Text (STT), is the task of transcribing a given audio to text. It has many applications, such as voice user interfaces.'}, 'relevance': 1}
Instruction: Given a `audio transcription` task, retrieve tools that process audio inputs to produce accurate textual transcriptions aligned with the query's requirements.

Query: Conduct a two-sample independent t-test with two samples, sample1=[1, 2, 3, 4, 5] and sample2=[6, 7, 8, 9, 10], and a significance level of 0.05.
Labels: [1] {'id': 'tool_id_693', 'doc': {'name': 'independent_samples_t_test', 'description': 'Conducts a two-sample independent t-test and returns the t-statistic, p-value, and conclusion.', 'parameters': {'sample1': {'description': 'The first sample of observations.', 'type': 'List[float]', 'default': 0.05}, 'sample2': {'description': 'The second sample of observations.', 'type': 'List[float]', 'default': 0.05}, 'alpha': {'description': 'The significance level of the test. Defaults to 0.05.', 'type': 'float, optional'}}}, 'relevance': 1}
Instruction: Given a `significance test` task, retrieve tools that perform statistical tests, specifically a two-sample independent t-test, by processing numerical inputs and returning the t-statistic, p-value.

Query: Can I get a list of all boards and their attributes on page number two with a page size of seven?
Labels: [1] {'id': 'ToolEyes_tool_34', 'doc': {'name': 'get_boards', 'description': 'A list of all boards and their attributes.', 'parameters': {'type': 'object', 'properties': {'page': {'type': 'string', 'description': 'Get the items on a specific page. 0(default) is the first page.'}, 'page_size': {'type': 'string', 'description': 'Get the number of boards on a specific page. Default: 5.'}}, 'required': []}}}
Instruction: Given a `pagination query` task, retrieve tools that can list boards and their attributes by processing parameters such as page number and page size to return the requested information.
\end{lstlisting}

\subsection{Human annotation}
To ensure the quality of the generated instructions, we conduct human annotation to review them based on four key aspects listed in Table~\ref{tab:quality}.  
For instructions deemed low-quality, human annotators manually revise them to improve accuracy and clarity.
For a clear illustration, we provide concrete examples of handcrafted instructions, high-quality generated instructions, and low-quality generated instructions in Table~\ref{tab:instruction-example}.
Below, we also provide the detailed human annotation guidelines used in our review process for reproducibility and transparency.

\begin{lstlisting}[basicstyle=\small\ttfamily, breaklines=true, breakindent=0em, commentstyle=\color{red!50!green!50!blue!50}, frame=shadowbox, rulesepcolor=\color{red!20!green!20!blue!20},numbers=none,literate={`}{\textasciigrave}1]
# Human guidance for instruction quality annotation

We ask you to evaluate the quality of the generated instructions based on the following four aspects. Please carefully assess each instruction and provide your judgment:

## Aspects for annotation

1. Hallucination Check: Does the instruction contain any incorrect or fabricated information about the target tools or the input query?  (Are there any details in the instruction that do not align with the actual features of the target tools or the content of the input query?)

2. Comprehensiveness of Tool Features: Does the instruction fully and accurately describe the features of all target tools mentioned in the query? (Are there any important features of the target tools that are missing or inadequately described in the instruction?)

3. Accuracy of Tool Feature Description: Does the instruction correctly describe the features of the target tools? (Key question to ask: Are the descriptions of the target tools technically accurate and consistent with their actual functionality?)

4. Relevance to Input Query: Is the instruction directly relevant to the original input query? (Key question to ask: Does the instruction address the specific needs or context provided in the input query, or does it deviate from the query's intent?)

## Detailed annotation Process

For each instruction, evaluate it based on the four aspects above. 
1. If the instruction meets all criteria (no hallucination, comprehensive, accurate, and relevant), mark it as correct.
2. If the instruction fails to meet any of the criteria, mark it as incorrect and provide a brief explanation of the issue (e.g., "contains hallucination," "missing key tool features," or "irrelevant to query").

For incorrect instructions, ``revise`` them to ensure they meet all quality criteria. The goal of this annotation process is to ensure that all instructions in our benchmark are of high quality and faithfully grounded in the original queries and target tools. 
\end{lstlisting}

\section{Large-scaling training dataset: \ours-train}

We extend the data collection process from \ours to incorporate the training sets of three mainstream tool-use datasets: ToolACE~\cite{toolace}, ToolBench~\cite{qin2023toolllm}, and APIGen~\cite{liu2024apigen}. These datasets are selected for their diversity in task types, tool categories, and query complexity, ensuring a comprehensive representation of real-world tool-use scenarios.

After collecting and preprocessing the data, we ultimately collect over 200k training instances. Each instance consists of a query 
 and a corresponding set of target tools.
Next, we further pair each query with an instruction using our target-aware strategy (\S~\ref{sec:data-instruction}). 
This strategy generates instructions that explicitly guide the retrieval process by incorporating task-specific context and tool functionality descriptions.
We report the basic statistics of \ours-train in Table~\ref{tab:training-set}.
We also show a training example of our \ours-train.

\begin{lstlisting}[basicstyle=\small\ttfamily, breaklines=true, breakindent=0em, commentstyle=\color{red!50!green!50!blue!50}, frame=shadowbox, rulesepcolor=\color{red!20!green!20!blue!20},numbers=none,literate={`}{\textasciigrave}1]

# Query: Is 'https://www.apple.com' available in the Wayback Machine on September 9, 2015?

# Instruction: Given a `URL availability` task, retrieve tools that check if a given URL is archived and accessible on a specific date in the Wayback Machine.

# Target tools (labels): ['{'name': 'availability', 'description': 'Checks if a given URL is archived and currently accessible in the Wayback Machine.', 'parameters': {'url': {'description': 'The URL to check for availability in the Wayback Machine.', 'type': 'str', 'de...}}}']

# Negative tools: [
    {'name': 'top_grossing_mac_apps', 'description': 'Fetches a list of the top-grossing Mac apps from the App Store.', 'parameters': {'category': {'description': "The category ID for the apps to be fetched. Defaults to '6016' (general category).", 'type': 'str', 'default': '6016'}, 'country': {'descript...}, 
    {'name': 'top_paid_mac_apps', 'description': 'Retrieves a list of the top paid Mac  apps from the App Store.', 'parameters': {'category': {'description'...},
    ... 
    {'name': 'exact_url_non_english', 'description': 'Retrieves the backlinks of a specific non-English URL using the RapidAPI service...}
]
\end{lstlisting}

\input{table/training-set}

\input{table/instruction}

\section{More experiment details}\label{sec:app:results}

\subsection{Baselines}
We comprehensively evaluate the following mainstream retrieval models on our benchmark, including:
\begin{itemize}
    \item \textbf{Sparse Retrieval}. These methods measure the similarity between tasks and tool documentation based on lexical overlap. We evaluate BM25s~\cite{L2024BM25SOO}.
    \item \textbf{Single-task dense retrieval}. These methods employ dual-encoder architecture models trained on conventional IR datasets. We evaluate gtr~\cite{Ni2021LargeDE}, contriever~\cite{Izacard2021UnsupervisedDI}, and colbertv2.0~\cite{colbertv2}, all trained on MS-MARCO~\cite{bajaj2016ms} with relevance criteria. We also evaluate the COLT~\cite{qu2024colt} which is a recently proposed model trained on an ad-hoc tool retrieval dataset.
    \item \textbf{Multi-task Embedding Models}. These methods utilize transformer encoders trained on various annotated IR datasets. We evaluate gte~\cite{Li2023TowardsGT}, bge~\cite{Xiao2023CPackPR}, and e5~\cite{wang2022text}, covering a wide range of parameter sizes. Additionally, we evaluate all-MiniLM-L6-v2\footnote{\href{https://huggingface.co/sentence-transformers/all-MiniLM-L6-v2}{huggingface.co/all-MiniLM-L6-v2}} from the Sentence Transformers platform.
    \item  \textbf{Cross-encoder Re-rankers}. These models re-rank the initially retrieved documents based on the query-passage relevance using bidirectional or unidirectional transformers. We evaluate MonoT5-Base and three re-rankers trained on diverse tasks: (i) mxbai-rerank-large-v1\footnote{\href{https://huggingface.co/mixedbread-ai/mxbai-rerank-large-v1}{huggingface.co/mxbai-rerank-large-v1}}, (ii) jina-reranker-v2-base\footnote{\href{https://huggingface.co/jinaai/jina-reranker-v2-base-multilingual}{huggingface.co/jina-reranker-v2-base-multilingual}}, and (iii) BGE-reranker.
    \item  \textbf{LLM Agents}. These methods leverage general-purpose LLM agents for re-ranking tasks in a zero-shot setting, simulating the tool selection process of tool-use agents. We evaluate the widely used LLM re-ranking framework, i.e., RankGPT~\cite{sun2023chatgpt}, with various LLMs as backbone.
\end{itemize}

We highlight that the initial tools for LLM agent and Re-ranking baselines are retrieved by \code{NV-embedd-v1 model}.
Details about these baselines are provided in Table~\ref{tab:baseline}.

\input{table/baseline}


\input{table/sep-with-inst}

\input{table/sep-wo-inst}

\begin{figure}[!t]
 \centering
\includegraphics[width=1\columnwidth]{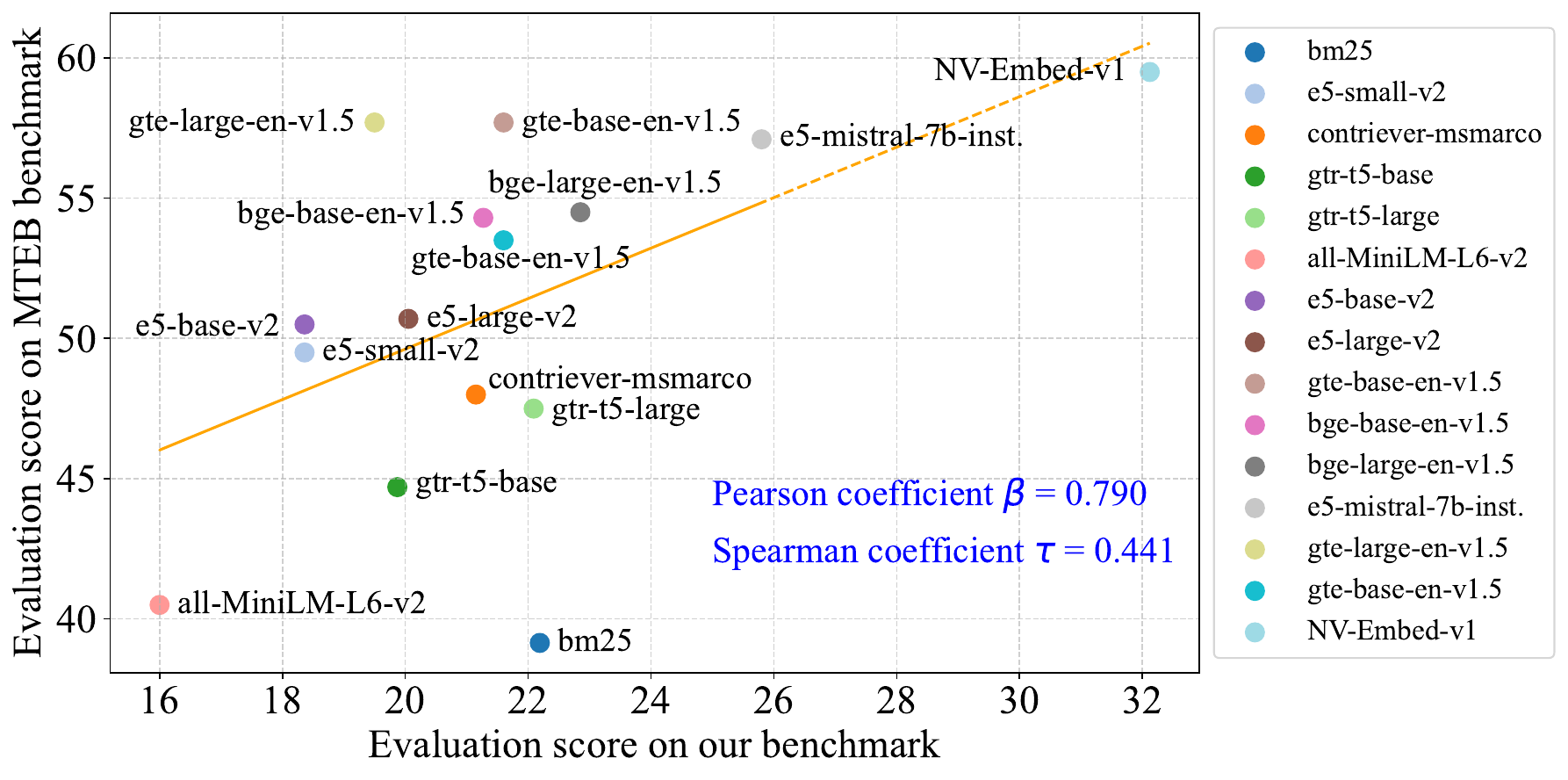} 
\caption{Correlation between the score on our benchmark and MTEB (retrieval subset).}
\label{fig:correlation-big}
\end{figure}

\subsection{Compare with conventional IR tasks}

To further investigate the complexity of tool retrieval tasks, we conducted a comparative analysis of model performance between our proposed benchmark (\ours) and the conventional Information Retrieval (IR) task benchmark, specifically the Massive Text Embedding Benchmark (MTEB). The relationship between these two benchmarks is visually presented in Figure~\ref{fig:correlation-big}.
Our analysis reveals two significant findings. 

\paragraph{First}, we observe a strong positive correlation between the two benchmarks, as evidenced by a Pearson's correlation coefficient of $\beta = 0.790$, indicating a similar performance trend across models. However, we observe that the absolute performance scores in \ours are consistently lower than those in the conventional IR benchmark. This discrepancy suggests that while our benchmark shares fundamental characteristics with conventional IR tasks, it presents additional challenges that make it more demanding for existing models.

\paragraph{Second}, our experimental results demonstrate that state-of-the-art IR models, particularly those trained with relevance-oriented optimization criteria (e.g., Contriever), exhibit substantially degraded performance on \ours. This performance gap underscores the necessity for target-aware reasoning capabilities in our benchmark, which goes beyond traditional relevance matching. The unique challenges of \ours are further elaborated in \S~\ref{sec:complexity}, where we identify two key distinguishing factors: (1) the presence of multiple potential target tools for each query, and (2) significantly lower term overlap between input queries and relevant tools compared to conventional IR scenarios. These characteristics collectively contribute to a more complex retrieval environment that requires advanced reasoning and understanding capabilities from retrieval models.

\subsection{Results of controlled experiment}
Since \ours integrates multiple datasets, we also conduct controlled experiments where IR models retrieve tools exclusively within the toolset of each individual dataset instead of the overall tool corpus.
Table~\ref{table:sep-wo-inst} presents the results under the setting that the IR models only take the query to retrieve, i.e., the \textit{w/o inst} setting.
Table~\ref{table:sep-w-inst} presents the results under the setting that the IR models take the query and additional instruction to retrieve, i.e., the \textit{w/ inst} setting.

\subsection{Results of in-subset retrieval}
\ours contains three subsets, including \ours-web, \ours-code and \ours-customized.
The tool in each subset diverges by its documentation format, domain, and functionality.
For a comprehensive evaluation, we also conduct an in-subset retrieval experiment, where  IR models retrieve tools exclusively within the toolset of each subset instead of the overall tool corpus.
Table~\ref{table:category-wo-inst} presents the results under the setting that the IR models only take the query to retrieve, i.e., the \textit{w/o inst} setting.
Table~\ref{table:category-w-inst} presents the results under the setting that the IR models take the query and additional instruction to retrieve, i.e., the \textit{w/ inst} setting.

\input{table/category-with-inst}
\input{table/category-without-inst}

\input{table/train}
\subsection{Results of trained IR mdels}
Experimental results on \ours reveal that even IR models with strong performance on conventional IR benchmarks such as MTEB and BEIR struggle significantly in tool retrieval tasks.  
A key factor contributing to this performance degradation is the lack of a large-scale training dataset specifically tailored for tool retrieval.
To address this gap, we introduce \ours-train, a diverse training dataset comprising more than 200k tool retrieval tasks.  
Each example in \ours-train consists of an input query, an instruction generated using our target-aware strategy, the corresponding target tools, and a set of negative tools.  
IR models are trained to distinguish target tools from negative tools (\S~\ref{sec:train}).  
We evaluate these trained IR models on \ours and present the results in Table~\ref{table:train}.

\input{table/pass-rate}
\subsection{Improved IR enhances tool-use LLMs}
We further investigate the impact of improved IR models on the end-to-end performance of tool-use LLMs.  
Specifically, we evaluate tool-use LLMs on the ToolBench~\cite{qin2023toolllm} dataset using the official \textit{Pass Rate} metric, which measures whether the model successfully invokes the correct tools to complete a given task.

For each task in ToolBench, we replace the pre-annotated toolset (oracle) with tools retrieved by IR models from \ours' tool corpus, which contains 43,000 tools.  
Since \ours integrates the ToolBench dataset, we can compute NDCG@10 for this retrieval step.
For a comprehensive evaluation, we assess two widely used tool-use LLMs, including GPT-3.5 and ToolLLaMA~\cite{qin2023toolllm}.

Table~\ref{tab:pass-rate} presents the retrieval NDCG@10 scores alongside the corresponding pass rates on ToolBench.\footnote{ToolBench consists of three subsets: ToolBench-G1, ToolBench-G2, and ToolBench-G3.}  
Our results demonstrate that LLM agents equipped with improved IR models achieve substantial gains in pass rate, highlighting the critical role of accurate tool retrieval in downstream task performance.
Furthermore, Figure~\ref{fig:pass} visually illustrates a positive correlation between improved IR performance and higher task pass rate, suggesting that better retrieval directly leads to improved downstream outcomes.

Based on this analysis, we propose that future work could explore the following two directions:
(i) Further optimize IR models to enhance tool retrieval performance; or
(ii) Adapt IR models by incorporating feedback from end-to-end task performance, allowing them to better support tool-use LLMs.  
These approaches provide a more efficient plug-and-play solution compared to fine-tuning LLMs, enabling flexible integration into diverse tool-use systems.






%% file: table/dataset.tex
\begin{table*}[!t]
\centering
\begin{adjustbox}{width=\columnwidth,center}
\begin{tabular}{@{} l l ccc @{}}
\toprule
\textbf{Dataset} & \textbf{Endpoint} & \textbf{Query size}  & \textbf{Tool size} & \textbf{Task type } \\
\rowcolor{Gainsboro}\multicolumn{5}{l}{\textit{Web APIs}} \\

GTA~\cite{wang2024gta} & \url{https://github.com/open-compass/GTA}  & 14 &  14 \\

Gorilla~\cite{patil2023gorilla}  &  \url{https://github.com/ShishirPatil/gorilla} & 598  & 1,005 \\
\textit{- gorilla-pytorch} subset & \url{https://github.com/ShishirPatil/gorilla/tree/main/data} &  43 &  43  \\
\textit{- gorilla-tensor } subset & \url{https://github.com/ShishirPatil/gorilla/tree/main/data} &  55 & 55 &  \\
\textit{- gorilla-huggingface} subset & \url{https://github.com/ShishirPatil/gorilla/tree/main/data} & 500 & 907 & \\

CRAFT~\cite{yuan2023craft} & \url{https://github.com/lifan-yuan/CRAFT} & 654 & 985  \\
\textit{- craft-Tabmwp} subset  & \url{https://github.com/lifan-yuan/CRAFT/tree/main/tab_and_math/TabMWP} & 174  & 180  \\
\textit{- craft-Vqa} subset  & \url{https://github.com/lifan-yuan/CRAFT/tree/main/vqa} &  200&  525\\
\textit{- craft-algebra} subset &  \url{https://github.com/lifan-yuan/CRAFT/tree/main/tab_and_math/MATH} & 280  & 280 \\

AutoTools~\cite{shi2024chain} & \url{https://github.com/mangopy/AutoTools} & 159 & 159  \\
\textit{- AutoTools-Food}  subset & \url{https://github.com/mangopy/AutoTools/tree/main/data} & 22 & 22 \\
\textit{- AutoTools-Weather} subset & \url{https://github.com/mangopy/AutoTools/tree/main/data} &   11  & 11 \\ 
\textit{- AutoTools-Movie} subset & \url{https://github.com/mangopy/AutoTools/tree/main/data} & 54  & 54 \\ 
\textit{- AutoTools-music} subset & \url{https://github.com/mangopy/AutoTools/tree/main/data} &  72 & 72 \\ 

APIGen~\cite{liu2024apigen} & \url{https://huggingface.co/datasets/Salesforce/xlam-function-calling-60k} & 1,000 & 3,605 \\
APIbank~\cite{li2023api}  &  \url{https://github.com/AlibabaResearch/DAMO-ConvAI} 
&  101 &  101 \\

Appbench~\cite{wang2024appbench}  &  \url{https://github.com/ruleGreen/AppBench} & 32 & 32\\

Mms~\cite{ma2024m} & \url{https://github.com/RAIVNLab/mnms}  & 33 &  33 \\

Metatool (\textit{a.k.a.}, ToolE)~\cite{huang2023metatool}  & \url{https://github.com/HowieHwong/MetaTool} & 200 & 200 \\

Reverse Chain~\cite{zhang2023reverse} & \url{https://github.com/ASK-03/Reverse-Chain} &  200 & 783 \\

RestGPT~\cite{song2023restgpt}   & \url{https://github.com/Yifan-Song793/RestGPT} &   94 & 94 \\
\textit{- RestGPT-TMDB} subset &  \href{https://github.com/Yifan-Song793/RestGPT/blob/main/datasets/tmdb.json}{https://github.com/Yifan-Song793/RestGPT/datasets/tmdb.json} & 54 & 54 \\ 
\textit{- RestGPT-Spotify} subset  & \href{https://github.com/Yifan-Song793/RestGPT/blob/main/datasets/spotify.json}{https://github.com/Yifan-Song793/RestGPT/datasets/spotify.json}  & 40 & 40 \\
  
Toolbench~\cite{qin2023toolllm} & \url{https://github.com/OpenBMB/ToolBench} &  1,100 & 13,862  \\
\textit{- G1-instruction} subset & \url{https://drive.google.com/drive/folders/1yBUQ732mPu-KclJnuQELEhtKakdXFc3J} & 200 & 13,862  \\
\textit{- G1-Tool} subset & \url{https://drive.google.com/drive/folders/1yBUQ732mPu-KclJnuQELEhtKakdXFc3J} &200 & 13,862 \\
\textit{- G1-category} subset & \url{https://drive.google.com/drive/folders/1yBUQ732mPu-KclJnuQELEhtKakdXFc3J} & 200  & 13,862 \\
\textit{- G2-instruction} subset & \url{https://drive.google.com/drive/folders/1yBUQ732mPu-KclJnuQELEhtKakdXFc3J} & 200 & 13,862  \\
\textit{- G2-instruction} subset & \url{https://drive.google.com/drive/folders/1yBUQ732mPu-KclJnuQELEhtKakdXFc3J} & 200  & 13,862  \\
\textit{- G3-instruction} subset & \url{https://drive.google.com/drive/folders/1yBUQ732mPu-KclJnuQELEhtKakdXFc3J} &100 & 13,862  \\

ToolLens~\cite{qu2024colt} & \url{https://github.com/quchangle1/COLT} & 314 & 314 \\
Tooleyes~\cite{ye2024tooleyes} & \url{https://github.com/Junjie-Ye/ToolEyes} & 95 & 95 \\
ToolACE~\cite{toolace} & \url{https://huggingface.co/datasets/Team-ACE/ToolACE/} &  1,000 & 16,072 \\
GPT4tools~\cite{yang2024gpt4tools} & \url{https://github.com/AILab-CVC/GPT4Tools} & 32 & 32  \\
Rotbench~\cite{ye2024rotbench} & \url{https://github.com/Junjie-Ye/RoTBench} & 550  & 919 \\
T-eval~\cite{chen2023t} & \url{https://github.com/open-compass/T-Eval} &  100 & 100 \\
\textit{- T-eval-step level} subset & \url{https://huggingface.co/datasets/lovesnowbest/T-Eval} &  50 & 50 \\
\textit{- T-eval-dialogue level} subset  & \url{https://huggingface.co/datasets/lovesnowbest/T-Eval} &  50 & 50 \\

Taskbench~\cite{shen2023taskbench} & \url{https://github.com/microsoft/JARVIS} 
& 103 & 103 \\
\textit{- TaskBench-multimedia} subset & \href{https://github.com/microsoft/JARVIS/tree/main/taskbench/data_multimedia}{https://github.com/microsoft/JARVIS/taskbench/multimedia} & 40 & 40\\
\textit{- TaskBench-daily} subset & \href{https://github.com/microsoft/JARVIS/tree/main/taskbench/data_dailylifeapis}{https://github.com/microsoft/JARVIS/taskbench/dailylifeapis}  & 40 & 40\\
\textit{- TaskBench-DL} subset & \href{https://github.com/microsoft/JARVIS/tree/main/taskbench/data_huggingface}{https://github.com/microsoft/JARVIS/taskbench/huggingface} & 23 & 23 \\

ToolAlpaca~\cite{tang2023toolalpaca} & \url{https://github.com/tangqiaoyu/ToolAlpaca} & 94 & 1,937 \\

Toolbench-sam~\cite{xu2023tool} & \url{https://github.com/sambanova/toolbench} & 197 & 197 \\

ToolEmu~\cite{ruan2023identifying} & \url{https://github.com/ryoungj/ToolEmu} & 38 & 38 \\

TooLink~\cite{qian2023toolink} & \url{https://github.com/qiancheng0/Toolink} & 497 & 1,804 \\

UltraTool~\cite{huang2024planning} & \url{https://github.com/JoeYing1019/UltraTool} & 500 &  1,171  \\
Tool-be-honest~\cite{zhang2024toolbehonestmultilevelhallucinationdiagnostic} & \url{https://github.com/ToolBeHonest/ToolBeHonest} & 350 & 892 \\
\bottomrule
\end{tabular}
\end{adjustbox}
\caption{The detailed statistics about the each collected dataset in \ours. We highlight that the subsets of ToolBench share a same toolsets containing 13,000+ tools. Besides, the \ours combine and deduplicate the toolsets from the above datasets to build the final tool retrieval corpus.}\label{tab:dataset}
\end{table*}

%% file: table/algorithm.tex
\RestyleAlgo{ruled}
\begin{algorithm}[!t]
\KwIn{A set of $N$ seed instructions $S = \{s_i \mid i \in [N]\}$ manually crafted by human experts; A powerful LLM $\mathcal{M}$ (e.g., GPT-4o); Collected tasks $\mathcal{T} = \{t_i | i \in [|\mathcal{T}|]\}$}
Initialize an instruction pool $I \gets S$\;
\For{$i$ $\in$ $|\mathcal{T}|$}{
    Sample $k$ examples $\{s_1', s_2', ..., s_k'\}$ from $I$\;
    \textcolor{dm-blue-500}{\texttt{//  Generate a new instruction $s_{i}$ using $\mathcal{M}$ through in-context learning:}} \\
    $s_i \gets \mathcal{M}(\text{prompt with } \{s_1', s_2', ..., s_k'\})$\;
    \textcolor{dm-purple-500}{\texttt{//Append new instruction to pool:}} \\ 
    $\mathcal{I}$ = $\mathcal{I} \cup \{s_i\}$ \;
}
Apply heuristic filtering to remove low-quality instructions from $I$\;
\KwOut{A set of high-quality instructions $\mathcal{I} = \{s_1, s_2, ..., s_{|\mathcal{T}|}\}$}
\caption{The pseudo algorithm for our target-aware strategy in automatically constructing instructions for evaluation tasks.}
\label{algo:instruction}
\end{algorithm}


%% file: table/case.tex


\begin{lstlisting}[basicstyle=\small\ttfamily, breaklines=true, breakindent=0em, commentstyle=\color{red!50!green!50!blue!50}, frame=shadowbox, rulesepcolor=\color{red!20!green!20!blue!20},numbers=none,literate={`}{\textasciigrave}1]
# An example of an evaluation task in our proposed benchmark

- Query: I need to find a grocery store near 123 Main Street, Downtown District that has a good selection of limes for my Easter celebration.
- ID: toolLens_query_7
- Target tools (labels): toolLens_tool_20, toolLens_tool_50, toolLens_tool_2
- Instruction: Given a `local grocery search` task, retrieve tools that can locate grocery stores based on the user's specified location and criteria, such as the availability of specific items like limes, to meet the query's requirements.

# Examples of tool documentation

- toolLens_tool_20: {"category_name": "Food", "required_parameters": [{"name": "ingredient", "type": "STRING", "description": "", "default": "strawberry"}], "optional_parameters": [], "method": "GET", "template_response": {"name": "str", "ingredients": ["list of str with length 9"], "instructions": ["list of str with length 7"]}, "name": "pastry/ingredient", "description": "This API endpoint allows users to retrieve a random pastry recipe that contains a specific ingredient. Users can make a GET request to the endpoint with the name of the ingredient as a query parameter, and the API will return a JSON response with the given recipe, including the name, list of ingredients, and instructions."}

- toolLens_tool_50: {"category_name": "Health_and_Fitness", "required_parameters": [], "optional_parameters": [{"name": "limit", "type": "NUMBER", "description": "limit the length of response", "default": "10"}], "method": "GET", "template_response": {"count": "int", "food": [{"_id": "str", "food_name": "str", "quantity": "str", "calories": "int", "uri": "str", "type": "str", "type_uri": "str", "core": "str", "core_uri": "str", "food_nutrition": [{"nutrient_name": "str", "value": "float", "unit": "str", "_list_length": 3}], "_list_length": 10}]}, "name": "View All Food Items", "description": "The request allows clients to retrieve a comprehensive list of all available food items.\n\nAPI request sent to [https://indnutrientsapi.tech/food](https://indnutrientsapi.tech/food)"}

- toolLens_tool_2: {"category_name": "Food", "required_parameters": [{"name": "grocery", "type": "string", "description": "", "default": ""}], "optional_parameters": [], "method": "GET", "template_response": {"message": "str"}, "name": "Search a Grocery", "description": "Search a specific grocery"}
\end{lstlisting}

%% file: table/training-set.tex
\begin{table}[h]
\centering
\begin{adjustbox}{width=0.8\columnwidth,center}
\begin{tabular}{@{}p{12cm} r@{}}
\toprule
\begin{tabular}[c]{@{}c@{}} \textbf{Statistic} \end{tabular}    
\\
\midrule
\# size of retrieval task &   205,826  \\
\# Average token length of the input query   & 52.87  \\
\# Average token length of the paired iinstruction  &  46.72 \\
\# Average token length of the tool documentation & 163.52 \\
\# Number of negative tools per input query &  5 \\
\# Number of target tools (labels)  per input query &  2.31 \\
\bottomrule
\end{tabular}
\end{adjustbox}
\caption{Basic statistics of the collected large-scaling training set \ours-train. We use the tokenizer from gpt-3.5-turbo in this work.}\label{tab:training-set}
\vspace{-2mm}
\end{table}

%% file: table/instruction.tex
\begin{table*}[htbp]
\centering
\begin{adjustbox}{width=\columnwidth,center}
\begin{tabular}{p{18cm}}
\hline

\rowcolor{Gainsboro} \multicolumn{1}{c}{\textit{Example of our seed instructions (handcrafted instruction}} \\

\# Query: I would like to generate a video presenting a text-based discussion on the topic of 'The Benefits of Exercise'. \\
\# Instruction: Given a "text-to-video" task, retrieve tools that process text inputs to generate coherent textual outputs aligned with the query's topic and requirements.\\
\midrule
\# Query: I have an audio file 'example.wav' which is difficult to understand. I would like you to help me transcribe the audio to text. \\
\# Instruction: Given a "audio transcription" task, retrieve tools that process audio inputs to produce accurate textual transcriptions aligned with the query's requirements.\\
\midrule
\# Query: Conduct a two-sample independent t-test with two samples, sample1=[1, 2, 3, 4, 5] and sample2=[6, 7, 8, 9, 10], and a significance level of 0.05. \\
\# Instruction: Given a "significance test" task, retrieve tools that perform statistical tests, specifically a two-sample independent t-test, by processing numerical inputs and returning the t-statistic, p-value.\\
\midrule
\# Query: Can you cancel a timer for my smart device?\\
\# Instruction: Given a "timer cancellation" task, retrieve tools that handle smart device operations by processing device ID and switch time inputs to cancel a scheduled action and return the status of the operation.\\
\midrule
\# Query: Find cruise tickets from Fontana to Santa Rosa on date 2023-07-04.\\
\# Instruction: Given a "ticket booking" task, retrieve tools that support booking cruise tickets by processing travel details such as departure location, destination, date, and time.\\

\hline
\rowcolor{Gainsboro} \multicolumn{1}{c}{\textit{Example of our generated instructions (high-quality instructions)}} \\

\# Query: Suppose that $f(x)=4x+5$. What is $f^{-1}(f^{-1}(9))$?\\
\# Instruction: Given a "inverse function calculation" task, retrieve tools that calculate the value of the repeated inverse for a linear function by processing coefficients, constants, and target values to determine the  result.\\

\midrule
\# Query: Identify an function that can classify images and works with spiking neural networks.\\
\# Instruction: Given an "image classification" task, retrieve tools that execute image classification by using spiking neural network models and processing image inputs.\\

\midrule

\# Query: I need to find a grocery store near 123 Main Street, Downtown District that has a good selection of limes for my Easter celebration.\\
\# Instruction": "Given a "local grocery search" task, retrieve tools that can locate grocery stores based on the user's specified location and criteria, such as the availability of specific items like limes, to meet the query's requirements.\\

\midrule

\# Query: Please help me find a recipe with no more than 40 grams of carbohydrates per gram and at least 5 grams of protein per gram. \\
\# Instruction: "Given a "nutritional recipe search" task, retrieve tools that can find recipes based on specific nutritional criteria such as carbohydrate and protein content.\\

\midrule

\# Query: What is 64277 times 38142? \\
\# Instruction: Given a "multiplication" task, retrieve tools that compute the product of two numbers by processing numerical inputs and returning the result. \\

\midrule
\# Query: Can I get a list of all boards and their attributes on page number two with a page size of seven? \\
\# Instruction: Given a "pagination query" task, retrieve tools that can list boards and their attributes by processing parameters such as page number and page size to return the requested information. \\

\hline
\rowcolor{Gainsboro} \multicolumn{1}{c}{\textit{Example of our generated instructions (low-quality instructions)}} \\
\# Query: I would like to generate a video presenting a text-based discussion on the topic of 'The Benefits of Exercise' \\
\# Instruction: Given a text-to-video task, please retrieve relevant tools to generate video about exercise.
\besttext{\textbf{// Too general to cover the key features.}} \\
\# Revised version: Given a `text-to-video` task,   retrieve tools related to general video scripting, exercise video libraries or tools process text data. \hightext{\textbf{// More specific to the query and target tools.}}\\
\midrule
\# Query: Can you assist me in finding a 1-bedroom townhouse or condo in Little Rock with max rent 1541000? I want it on the sixth floor with 7 balconies.\\
\# Instruction: Please retrieve tools find 1-bedroom or condo in Little Rock. \besttext{\textbf{// Too general and miss the point of \textit{floor}}} \\
\# Revised version: Given a property search task, retrieve tools that can find rental properties based on location, property type, rent budget, and specific features or requirements. \hightext{\textbf{// Cover all key points and related to tools functionality.}}\\
\bottomrule
\end{tabular}
\end{adjustbox}
\caption{Example of the instruction in \ours. We show the handcrafted instruction by human experts, the high-quality instruction and low-quality instruction generated by GPT-4o, respectively.
We also show the \hightext{revised version} of the low-quality instruction paired with \besttext{the reason for revision}.} \label{tab:instruction-example}
\end{table*}

%% file: table/baseline.tex
\begin{table*}[!t]
\centering
\begin{adjustbox}{width=0.95\columnwidth,center}
\begin{tabular}{@{} l l @{}}
\toprule
\textbf{Name} & \textbf{Public Link of Endpoint}\\ 
\hline

\rowcolor{Gainsboro}\multicolumn{2}{l}{\textit{Conventional sparse and dense models}} \\
BM25S~\cite{L2024BM25SOO} & \url{https://github.com/xhluca/bm25s} \\
contriever~\cite{izacard2021unsupervised} &  \url{https://huggingface.co/facebook/contriever-msmarco} \\
ColBERTv2~\cite{santhanam2021colbertv2} & \url{https://github.com/stanford-futuredata/ColBERT} \\
gtr-t5-base~\cite{ni2021large} & \url{https://huggingface.co/sentence-transformers/gtr-t5-base} \\
gtr-t5-large~\cite{ni2021large} & \url{https://huggingface.co/sentence-transformers/gtr-t5-large} \\
\hline
\rowcolor{Gainsboro}\multicolumn{2}{l}{\textit{Multi-task embedding models}} \\
all-MiniLM-L6-v2 & \url{https://huggingface.co/sentence-transformers/all-MiniLM-L6-v2} \\
e5-small-v2~\cite{wang2022text} & \url{https://huggingface.co/intfloat/e5-small-v2} \\
e5-base-v2~\cite{wang2022text} & \url{https://huggingface.co/intfloat/e5-base-v2} \\
e5-large-v2~\cite{wang2022text} & \url{https://huggingface.co/intfloat/e5-large-v2} \\ 
bge-base-en-v1.5~\cite{bge_embedding} & \url{https://huggingface.co/BAAI/bge-base-en-v1.5} \\
bge-large-en-v1.5~\cite{bge_embedding} & \url{https://huggingface.co/BAAI/bge-large-en-v1.5} \\
\midrule
gte-Qwen2-1.5B-inst.~\cite{Li2023TowardsGT} & \url{https://huggingface.co/Alibaba-NLP/gte-Qwen2-1.5B-instruct} \\
e5-mistral-7b~\cite{wang2023improving} & \url{https://huggingface.co/intfloat/e5-mistral-7b-instruct} \\
GritLM-7B~\cite{muennighoff2024generative} & \url{https://huggingface.co/GritLM/GritLM-7B} \\
NV-Embed-v1~\cite{lee2024nv} & \url{https://huggingface.co/nvidia/NV-Embed-v1} \\
\hline
\rowcolor{Gainsboro}\multicolumn{2}{l}{\textit{Cross-encoder re-ranking}} \\
mxbai-rerank-large-v1 & \url{https://huggingface.co/mixedbread-ai/mxbai-rerank-large-v1} \\
monot5-base~\cite{nogueira2020document} & \url{https://huggingface.co/castorini/monot5-base-med-msmarco} \\
bge-reranker-v2-m3~\cite{li2023making,chen2024bge} & \url{https://huggingface.co/BAAI/bge-reranker-v2-m3} \\
jina-reranker-v2 & \url{https://huggingface.co/jinaai/jina-reranker-v2-base-multilingual} \\
bge-reranker-v2-gemma~\cite{li2023making,chen2024bge} & \url{https://huggingface.co/BAAI/bge-reranker-v2-gemma} \\
\hline
\rowcolor{Gainsboro}\multicolumn{2}{l}{\textit{LLM agent}} \\
RankGPT & \url{https://github.com/sunnweiwei/RankGPT} \\
\textit{ -} Mixtral-8x22B & \url{https://huggingface.co/mistralai/Mixtral-8x22B-Instruct-v0.1} \\
\textit{ -} GPT-3.5-turbo-1106 & \url{https://openai.com/chatgpt/overview/}  \\ 
\textit{ -} GPT-3.5-turbo-0125 &  \url{https://openai.com/chatgpt/overview/} \\
 
\bottomrule
\end{tabular}
\end{adjustbox}
\caption{The public link or endpoint of the baselines in our experiments.}\label{tab:baseline}
\label{tab:main}
\end{table*}

%% file: table/sep-with-inst.tex
\begin{table*}[ht]
\centering
\begin{adjustbox}{width=1\columnwidth,center}
\setlength\tabcolsep{4pt}
\begin{tabular}{l cccc cccc cccc |cc}

\toprule
\multirow{2}{*}{\textbf{Model}} & 
\multicolumn{4}{c}{\textbf{REST API}} & 
\multicolumn{4}{c}{\textbf{Code Function}} & 
\multicolumn{4}{c}{\textbf{Customized tool}} & 
\multicolumn{2}{c}{\textbf{Avg.}} \\
\cmidrule(lr){2-5} \cmidrule(lr){6-9} \cmidrule(lr){10-13}  \cmidrule(lr){14-15}
& N@10 & P@10 & R@10 & C@10
& N@10 & P@10 & R@10 & C@10
& N@10 & P@10 & R@10 & C@10
& N@10 & C@10
\\
\hline
\rowcolor{Gainsboro}\multicolumn{15}{l}{\textit{Conventional sparse and dense models}} \\
\code{BM25s}  &  \high{62.09}  &  \high{15.68}  &  \high{72.98}  &  \high{58.06}  &  \high{56.98}  &  \high{8.24}  &  \high{73.95}  &  \high{72.81}  &  68.48  &  \high{14.78}  &  \high{80.51}  &  69.55  &  \high{62.51} &  \high{66.81} \\
\code{ColBERT}  &  56.73  &  14.59  &  67.86  &  47.87  &  53.56  &  7.82  &  71.66  &  70.09  &  64.49  &  13.05  &  76.35  &  64.85  &  58.26  &  60.94\\
\code{contriever-msmarco}  &  57.81  &  15.33  &  70.77  &  50.87  &  49.66  &  7.29  &  65.99  &  64.56  &  \high{68.86}  &  14.67  &  83.05  &  73.45  &  58.77  &  62.96\\
\code{gtr-t5-base}  &  57.06  &  14.54  &  68.26  &  49.75  &  49.38  &  7.29  &  65.76  &  64.09  &  70.48  &  14.29  &  81.60  &  70.96  &  58.97  &  61.60\\
\code{gtr-t5-large}  &  60.32  &  15.27  &  72.05  &  54.03  &  52.79  &  7.41  &  67.28  &  65.79  &  72.03  &  14.69  &  84.39  &  \high{73.95}  &  61.72  &  64.59 \\

\hline
\rowcolor{Gainsboro}\multicolumn{15}{l}{\textit{Embedding models}} \\

\code{all-MiniLM-L6-v2}  &  53.92  &  14.47  &  66.77  &  48.64  &  50.14  &  7.58  &  68.39  &  66.74  &  68.31  &  14.20  &  81.57  &  71.14  &  57.46  &  62.17\\
\code{e5-small-v2}  &  61.95  &  15.84  &  72.91  &  53.34  &  51.45  &  7.76  &  68.16  &  65.46  &  69.13  &  14.24  &  79.69  &  68.33  &  60.85  &  62.38\\
\code{e5-base-v2}  &  62.90  &  15.90  &  73.98  &  53.83  &  55.81  &  8.44  &  74.17  &  72.53  &  69.96  &  14.98  &  \high{83.63}  &  73.94  &  62.89  &  \high{66.77} \\
\code{e5-large-v2}  &  61.72  &  15.90  &  73.27  &  52.84  &  56.21  &  8.42  &  75.25  &  \high{73.14}  &  69.88  &  \high{15.01}  &  81.13  &  \high{71.30}  &  62.60  &  65.76\\
\code{gte-base-en-v1.5}  &  64.35  & \high{16.55} &  75.80  &  57.38  &  \high{59.18}  &  \high{8.77}  &  \high{76.95}  &  \high{74.45}  &  71.79  &  14.53  &  81.90  &  70.07  &  65.11  &  67.30 \\
\code{gte-large-en-v1.5}  &  60.67  &  15.46  &  72.30  &  52.41  &  54.11  &  8.22  &  73.35  &  71.37  &  68.59  &  14.36  &  80.41  &  69.82  &  61.12  &  64.53\\
\code{bge-base-en-v1.5}  &  65.05  &  16.37  &  75.72  &  57.30  &  54.55  &  7.72  &  69.22  &  67.48  &  71.21  &  14.71  &  83.13  &  72.53  &  63.60  &  65.77\\
\code{bge-large-en-v1.5}  & \high{66.25}  &  16.48  &  \high{75.84}  &  \high{57.75}  &  58.61  &  8.41  &  74.91  &  72.74  &  \high{71.19}  &  14.20  &  80.44  &  69.27  &  \high{65.35}  &  66.59 \\
\midrule
\code{gte-Qwen2-1.5B-inst.}  &  67.57  &  16.93  &  78.14  &  60.81  &  58.12  &  8.51  &  75.41  &  73.39  &  71.73  &  15.34  &  83.03  &  73.39  &  65.81  &  69.19\\
\code{e5-mistral-7b}  &  \high{69.51}  &  \high{17.37}  &  \high{79.34}  &  \high{62.48}  &  58.15  &  8.37  &  75.12  &  72.79  &  72.52  &  14.68  &  81.79  &  71.49  &  66.73  &  68.92\\
\code{GritLM-7B}  &  69.43  &  17.25  &  78.97  &  61.67  &  62.78  &  9.22  &  78.74  &  77.59  &  \high{76.04}  &  15.44  &  85.55  &  74.35  &  \high{69.42}  &  71.21\\
\code{NV-Embed-v1}  &  66.04  &  16.88  &  77.19  &  59.06  &  \high{63.46}  &  \high{9.40}  &  \high{81.79} &  \high{79.82}  &  75.39  &  \high{15.75}  &  \high{88.48}  &  \high{78.37}  &  68.30  &  \high{72.42} \\

\hline
\rowcolor{Gainsboro}\multicolumn{15}{l}{\textit{Cross-encoder re-ranking models}} \\
\code{mxbai-rerank-large-v1}  &  57.48  &  14.60  &  68.65  &  49.54  &  50.37  &  7.75  &  69.59  &  67.88  &  62.24  &  13.32  &  73.26  &  61.24  &  56.70  &  59.55\\
\code{monot5-base-msmarco}  &  54.57  &  14.23  &  64.38  &  46.12  &  50.00  &  8.05  &  68.76  &  66.80  &  64.50  &  13.28  &  75.80  &  67.84  &  56.36  &  60.25\\
\code{bge-reranker-v2-m3}  &  70.42  &  17.75  &  80.33  &  65.49  &  64.22  &  9.37  &  80.60  &  79.48  &  75.70  &  16.15  &  88.87  &  78.65  &  70.11  &  74.54\\
\code{bge-reranker-v2-gemma}  & \high{75.67}  &  \high{18.63}  &  \high{84.07}  &  \high{71.00} &  \high{69.59}  &  \high{9.78}  &  \high{84.18}  &  \high{83.47}  &  \high{77.17}  &  \high{16.55}  &  \high{88.34}  &  \high{79.80}  &  \high{74.14}  &  \high{78.09} \\

\bottomrule
\end{tabular}
\end{adjustbox}
\caption{Results of control experiment where each IR models is evaluated from the toolset of each integrated dataset in \textbf{\textit{w/ inst.}} setting.}
\label{table:sep-w-inst}
\end{table*}

%% file: table/sep-wo-inst.tex
\begin{table*}[ht]
\centering
\begin{adjustbox}{width=1\columnwidth,center}
\setlength\tabcolsep{4pt}
\begin{tabular}{l cccc cccc cccc cc}

\toprule
\multirow{2}{*}{\textbf{Model}} & 
\multicolumn{4}{c}{\textbf{\ours-Web}} & 
\multicolumn{4}{c}{\textbf{\ours-Code}} & 
\multicolumn{4}{c}{\textbf{\ours-Customized}} & 
\multicolumn{2}{c}{\textbf{Avg.}} \\
\cmidrule(lr){2-5} \cmidrule(lr){6-9} \cmidrule(lr){10-13}  \cmidrule(lr){14-15}
& N@10 & P@10 & R@10 & C@10
& N@10 & P@10 & R@10 & C@10
& N@10 & P@10 & R@10 & C@10
& N@10 & C@10
\\
\hline
\rowcolor{Gainsboro}\multicolumn{15}{l}{\textit{Conventional sparse and dense models}} \\
\code{BM25S}  &  51.79  &  13.78  &  63.35  &  45.46  &  \high{38.74}  &  \high{5.87}  &  52.65  &  51.39  &  59.72  &  \high{13.55}  &  \high{71.83}  &  60.38  &  50.08  &  52.41 \\
\code{ColBERT}  &  51.70  &  13.56  &  61.92  &  41.01  &  38.60  &  6.05  &  \high{55.07}  &  \high{54.05}  &  53.91  &  12.10  &  66.85  &  55.29  &  48.07  &  50.11 \\
\code{contriever-msmarco}  &  53.23  &  14.55  &  65.96  &  46.59  &  35.97  &  5.79  &  52.32  &  50.84  &  \high{56.94}  &  13.21  &  72.11  &  61.17  &  48.71  &  52.87\\
\code{gtr-t5-base}  &  51.65  &  14.13  &  64.55  &  44.56  &  33.98  &  5.51  &  48.88  &  47.77  &  54.28  &  13.24  &  70.95  &  60.82  &  46.64  &  51.05\\
\code{gtr-t5-large}  &  \high{56.62}  &  \high{15.06}  &  \high{69.77}  &  \high{50.26}  &  37.40  &  5.85  &  52.41  &  51.27  &  56.24  &  13.31  &  71.25  &  \high{61.40}  &  \high{50.09}  &  \high{54.31} \\

\hline
\rowcolor{Gainsboro}\multicolumn{15}{l}{\textit{Embedding models}} \\
\code{all-MiniLM-L6-v2}  &  48.49  &  13.54  &  62.35  &  43.63  &  34.40  &  5.62  &  50.15  &  48.71  &  58.08  &  13.22  &  72.72  &  62.17  &  46.99  &  51.50 \\
\code{e5-small-v2}  &  54.40  &  14.60  &  66.47  &  46.76  &  35.18  &  5.67  &  50.70  &  49.19  &  56.85  &  13.45  &  73.43  &  60.86  &  48.81  &  52.27\\
\code{e5-base-v2}  &  55.42  &  14.92  &  67.90  &  48.10  &  38.35  &  6.23  &  56.08  &  54.90  &  59.96  &  14.18  &  76.18  &  66.55  &  51.24  &  56.52\\
\code{e5-large-v2}  &  54.32  &  14.81  &  67.69  &  47.89  &  40.24  &  6.23  &  56.33  &  54.85  &  59.40  &  13.79  &  72.96  &  60.37  &  51.32  &  54.37 \\
\code{gte-base-en-v1.5}  &  56.48  &  \high{15.40}  &  \high{70.07}  &  \high{50.96}  &  39.46  &  \high{6.27}  &  \high{56.58}  &  \high{55.24}  &  \high{64.00}  &  \high{14.23}  &  \high{78.03}  &  \high{66.93}  &  53.31  &  \high{57.71} \\
\code{gte-large-en-v1.5}  &  55.39  &  14.89  &  68.33  &  48.98  &  38.23  &  6.22  &  56.16  &  54.95  &  57.88  &  13.86  &  75.12  &  65.15  &  50.50  &  56.36 \\
\code{bge-base-en-v1.5}  &  56.17  &  14.86  &  68.30  &  49.05  &  38.71  &  6.05  &  54.35  &  53.08  &  59.40  &  13.93  &  75.38  &  64.81  &  51.43  &  55.65 \\ 
\code{bge-large-en-v1.5}  &  \high{58.20}  &  15.33  &  69.83  &  50.20  &  \high{40.39}  &  6.13  &  55.03  &  53.67  &  61.40  &  13.83  &  77.27  &  66.71  &  \high{53.33}  &  56.86 \\ 
\midrule
\code{gte-Qwen2-1.5B-inst.}$\spadesuit$  &  60.28  &  15.64  &  72.60  &  53.82  &  44.06  &  6.56  &  59.29  &  57.78  &  65.57  &  14.79  &  \high{80.33}  &  \high{70.59}  &  56.64  &  60.73 \\
\code{e5-mistral-7b}$\spadesuit$  &  60.78  &  15.93  &  73.12  &  55.60  &  44.20  &  6.77  &  61.18  &  59.79  &  60.56  &  13.80  &  74.47  &  63.38  &  55.18  &  59.59\\
\code{GritLM-7B}$\spadesuit$  &  \high{62.54}  &  \high{16.07}  &  73.82  &  54.46  &  46.80  &  6.98  &  62.89  &  61.35  &  \high{67.61}  &  \high{14.93}  &  80.04  &  68.41  &  58.98  &  61.41 \\
\code{NV-Embed-v1}$\spadesuit$  &  61.76  &  16.02  &  \high{73.86}  &  \high{55.96}  &  \high{50.38}  &  \high{7.54}  &  \high{67.93}  &  \high{66.03}  &  67.01  &  14.61  &  79.70  &  69.74  &  \high{59.72}  &  \high{63.91} \\

\hline
\rowcolor{Gainsboro}\multicolumn{15}{l}{\textit{Cross-encoder re-ranking models}} \\
\code{mxbai-rerank-large-v1}  &  56.45  &  14.51  &  67.51  &  49.02  &  42.55  &  6.28  &  57.85  &  55.96  &  54.63  &  13.03  &  72.68  &  60.20  &  51.21  &  55.06 \\
\code{monot5-base-msmarco}  &  56.10  &  14.82  &  65.29  &  47.12  &  41.05  &  6.31  &  57.20  &  55.14  &  64.60  &  13.91  &  75.79  &  65.76  &  53.92  &  56.01\\
\code{bge-reranker-v2-m3}  &  61.78  &  15.94  &  72.35  &  55.61  &  45.15  &  6.74  &  61.03  &  59.56  &  62.45  &  14.86  &  79.65  &  68.68  &  56.46  &  61.28 \\
\code{jina-reranker-v2-base}  &  \high{65.86}  &  \high{17.14}  &  \high{77.54}  &  \high{62.33}  &  47.23  &  7.01  &  63.50  &  62.24  &  \high{69.10}  &  15.38  &  81.29  &  69.63  &  60.73  &  64.73 \\
\code{bge-reranker-v2-gemma}  &  65.80  &  16.87  &  76.85  &  61.40  &  \high{52.49}  &  \high{7.60}  &  \high{68.14}  &  \high{65.94}  &  67.87  &  \high{15.63}  &  \high{81.98}  &  \high{72.24}  &  \high{62.05}  &  \high{66.53} \\


\bottomrule
\end{tabular}
\end{adjustbox}
\caption{Results of control experiment where each IR models is evaluated from the toolset of each integrated dataset in \textbf{\textit{w/o inst.}} setting.}
\label{table:sep-wo-inst}
\end{table*}

%% file: table/category-with-inst.tex
\begin{table*}[ht]
\centering
\setlength\tabcolsep{4pt}
\resizebox{\textwidth}{!}{%
\begin{tabular}{l cccc cccc cccc cc}
\toprule
\multirow{2}{*}{\textbf{Model}} & 
\multicolumn{4}{c}{\textbf{\ours-Web}} & 
\multicolumn{4}{c}{\textbf{\ours-Code}} & 
\multicolumn{4}{c}{\textbf{\ours-Customized}} & 
\multicolumn{2}{c}{\textbf{Average}} \\
\cmidrule(l){2-15}
& N@10 & P@10 & R@10 & C@10
& N@10 & P@10 & R@10 & C@10
& N@10 & P@10 & R@10 & C@10
& N@10 & C@10 \\
\hline
\rowcolor{Gainsboro}\multicolumn{15}{l}{\textit{Sparse and dense models}} \\
\code{bm25}  &  \high{49.01}  &  \high{7.17}  &  \high{64.96}  &  \high{63.68} &  \high{28.92}  &  \high{6.78}  &  \high{37.09}  &  \high{24.44}  &  51.28  &  10.66  &  60.70  &  48.40  &  \high{43.07}  &  \high{45.51} \\
\code{COLT}  &  36.58  &  5.74  &  50.84  &  49.04  &  21.98  &  5.12  &  29.68  &  20.03  &  46.02  &  9.12  &  58.02  &  45.27  &  34.86  &  38.12\\
\code{Colbert}  &  43.80  &  6.40  &  58.02  &  56.28  &  16.60  &  3.05  &  20.85  &  14.95  &  31.18  &  5.86  &  39.40  &  32.10  &  30.53  &  34.44\\
\code{contriever-msmarco}  &  35.78  &  5.31  &  47.17  &  46.07  &  25.19  &  5.67  &  31.95  &  20.56  &  44.37  &  9.35  &  57.53  &  46.80  &  35.11  &  37.81\\
\code{gtr-t5-base}  &  37.45  &  5.50  &  48.78  &  47.44  &  22.54  &  4.93  &  29.64  &  20.60  &  51.02  &  10.28  &  61.08  &  49.06  &  37.00  &  39.03\\
\code{gtr-t5-large}  &  42.14  &  5.98  &  53.59  &  52.19  &  26.60  &  5.71  &  33.68  &  22.38  &  \high{53.95}  &  \high{11.17}  &  \high{66.08}  &  \high{52.21}  &  40.90  &  42.26\\

\hline
\rowcolor{Gainsboro} \multicolumn{15}{l}{\textit{Embedding models}}\\
\code{all-MiniLM-L6-v2}  &  36.93  &  5.65  &  49.54  &  47.92  &  15.89  &  3.89  &  22.68  &  15.24  &  43.09  &  9.29  &  56.84  &  43.96  &  31.97  &  35.71\\
\code{e5-small-v2}  &  38.22  &  5.55  &  49.34  &  48.07  &  28.97  &  6.81  &  36.96  &  23.45  &  47.59  &  9.78  &  58.19  &  45.31  &  38.26  &  38.94\\
\code{e5-base-v2}  &  40.69  &  6.51  &  55.38  &  54.01  &  28.43  &  6.59  &  37.14  &  23.67  &  47.89  &  9.48  &  59.01  &  46.91  &  39.00  &  41.53\\
\code{e5-large-v2}  &  40.14  &  5.87  &  52.23  &  50.80  &  26.88  &  6.16  &  35.65  &  24.31  &  51.40  &  10.65  &  61.45  &  48.28  &  39.47  &  41.13 \\
\code{gte-base-en-v1.5}  &  \high{48.25}  &  \high{7.16}  &  \high{61.96}  &  \high{59.17}  &  33.28  &  7.57  &  41.99  &  27.20  &  50.33  &  9.70  &  62.05  &  50.09  &  \high{43.95}  &  \high{45.49} \\
\code{gte-large-en-v1.5}$^\spadesuit$   &  40.48  &  6.46  &  56.34  &  54.52  &  30.58  &  7.00  &  38.79  &  24.78  &  49.24  &  9.98  &  59.13  &  47.20  &  40.10  &  42.17 \\
\code{bge-base-en-v1.5}  &  43.74  &  6.43  &  57.33  &  55.85  &  29.83  &  6.96  &  38.86  &  25.67  &  52.41  &  \high{10.75}  &  \high{63.84}  &  51.19  &  41.99  &  44.24 \\
\code{bge-large-en-v1.5}  &  44.07  &  6.44  &  56.78  &  55.22  &  \high{33.88}  &  \high{7.90}  &  \high{43.11}  &  \high{28.62}  &  \high{53.48}  &  10.53  &  63.66  &  \high{52.00}  &  43.81  &  45.28 \\
\midrule
\code{gte-Qwen2-1.5B-inst.}$^\spadesuit$   &  47.29  &  7.10  &  61.89  &  59.98  &  39.30  &  9.90  &  48.58  &  29.51  &  55.56  &  11.67  &  65.55  &  51.75  &  47.38  &  47.08\\
\code{e5-mistral-7b}$^\spadesuit$  &  48.76  &  7.16  &  63.57  &  61.40  &  33.06  &  8.14  &  43.56  &  28.49  &  57.16  &  11.43  &  67.28  &  52.62  &  46.32  &  47.51\\
\code{GritLM-7B}  &  \high{53.69}  &  \high{8.13}  &  \high{68.70}  &  \high{67.26}  &  \high{41.59}  &  \high{10.06}  &  \high{51.69}  &  \high{33.87}  &  \high{60.14}  &  11.63  &  68.93  &  54.60  &  \high{51.81}  &  \high{51.91} \\
\code{NV-Embed-v1}$^\spadesuit$   &  51.95  &  7.62  &  66.58  &  64.21  &  34.42  &  8.40  &  43.66  &  29.65  &  57.93  &  \high{12.34}  &  \high{71.14}  &  \high{57.47}  &  48.10  &  50.44 \\

\hline
\rowcolor{Gainsboro}\multicolumn{15}{l}{\textit{Cross-encoder re-ranking models}} \\
\code{mxbai-rerank-large-v1}  &  33.39  &  5.09  &  46.75  &  45.11  &  24.90  &  5.95  &  32.20  &  19.52  &  35.62  &  7.35  &  43.96  &  34.14  &  31.30  &  32.92\\
\code{monot5-base-msmarco}  &  29.95  &  5.20  &  42.47  &  40.36  &  30.20  &  7.84  &  37.94  &  21.43  &  46.99  &  9.26  &  57.10  &  46.48  &  35.71  &  36.09\\
\code{bge-reranker-v2-m3}  &  56.49  &  8.42  &  72.26  &  70.67  &  \high{38.09}  &  \high{9.25}  & \high{48.14}  &  \high{33.48}  &  54.66  &  12.46  &  70.79  &  56.17  &  49.75  &  53.44\\
\code{jina-reranker-v2-base}  &  35.24  &  5.48  &  47.70  &  46.31  &  36.23  &  9.21  &  45.59  &  29.26  &  54.12  &  11.92  &  65.09  &  51.18  &  41.86  &  42.25\\
\code{bge-reranker-v2-gemma}  &  \high{62.84}  & \high{8.87}  &  \high{76.24}  &  \high{74.86}  &  37.13  &  8.62  &  47.23  &  33.45  &  \high{64.33}  &  \high{13.72}  &  \high{77.37}  &  \high{62.08}  &  \high{54.76}  &  \high{56.79} \\
\bottomrule
\end{tabular}
}
\caption{Experiments are conducted under the \textbf{\textit{w/ inst.}} setting, with retrieval performed within each subset individually.}\label{table:category-w-inst}
\end{table*}

%% file: table/category-without-inst.tex
\begin{table*}[ht]
\centering
\setlength\tabcolsep{4pt}
\resizebox{\textwidth}{!}{%
\begin{tabular}{l cccc cccc cccc cc}
\toprule
\multirow{2}{*}{\textbf{Model}} & 
\multicolumn{4}{c}{\textbf{\ours-Web}} & 
\multicolumn{4}{c}{\textbf{\ours-Code}} & 
\multicolumn{4}{c}{\textbf{\ours-Customized}} & 
\multicolumn{2}{c}{\textbf{Average}} \\
\cmidrule(l){2-15}
& N@10 & P@10 & R@10 & C@10
& N@10 & P@10 & R@10 & C@10
& N@10 & P@10 & R@10 & C@10
& N@10 & C@10 \\
\hline
\rowcolor{Gainsboro}\multicolumn{15}{l}{\textit{Sparse and dense models}} \\
\code{bm25}  &  \high{26.47}  &  \high{4.04}  &  \high{34.51}  &  \high{33.22}  &  20.61  &  5.06  &  26.35  &  15.87  &  38.48  &  8.51  &  48.24  &  37.09  &  28.52  &  28.73\\
\code{COLT}  &  22.23  &  3.75  &  31.66  &  30.15  &  21.65  &  5.36  &  29.15  &  19.12  &  36.12  &  7.99  &  47.63  &  37.64  &  26.67  &  28.97\\
\code{Colbert}  &  23.58  &  3.84  &  32.64  &  31.19  &  \high{23.50}  &  5.70  &  28.82  &  16.03  &  33.71  &  6.64  &  40.85  &  32.01  &  26.93  &  26.41\\
\code{contriever-msmarco}  &  19.11  &  3.16  &  26.52  &  24.97  &  21.84  &  \high{5.93}  &  27.83  &  15.19  &  35.01  &  7.57  &  44.07  &  34.71  &  25.32  &  24.96\\
\code{gtr-t5-base}  &  21.19  &  3.48  &  29.05  &  27.90  &  18.18  &  4.45  &  25.09  &  16.32  &  35.95  &  8.10  &  45.77  &  36.50  &  25.11  &  26.91\\
\code{gtr-t5-large}  &  24.48  &  3.91  &  33.55  &  32.44  &  23.49  &  5.64  &  \high{30.79}  &  \high{19.31}  &  \high{38.88}  &  \high{8.89}  &  \high{49.55}  &  \high{38.64}  &  \high{28.95}  &  \high{30.13} \\

\hline
\rowcolor{Gainsboro} \multicolumn{15}{l}{\textit{Embedding models}}\\
\code{all-MiniLM-L6-v2}  &  18.07  &  3.10  &  25.13  &  23.75  &  13.71  &  3.49  &  18.73  &  11.78  &  31.61  &  7.12  &  40.66  &  30.03  &  21.13  &  21.85\\
\code{e5-small-v2}  &  20.10  &  3.14  &  26.47  &  25.12  &  21.02  &  5.33  &  27.70  &  16.88  &  32.58  &  7.35  &  39.85  &  30.33  &  24.57  &  24.11\\
\code{e5-base-v2}  &  20.96  &  3.54  &  29.43  &  28.23  &  21.25  &  5.43  &  27.59  &  16.42  &  32.90  &  7.33  &  43.10  &  33.51  &  25.04  &  26.05\\
\code{e5-large-v2}  &  22.93  &  3.49  &  29.53  &  28.09  &  20.16  &  5.19  &  27.26  &  16.68  &  \high{39.45}  &  \high{8.81}  &  49.07  &  38.19  &  27.51  &  27.65\\
\code{gte-base-en-v1.5}  &  \high{24.50}  &  \high{3.98}  &  \high{33.85}  & \high{32.48}  &  24.32  &  6.46  &  33.01  &  19.91  &  37.86  &  8.10  &  \high{49.44}  &  \high{38.94}  &  \high{28.89}  &  \high{30.44} \\
\code{gte-large-en-v1.5}$^\spadesuit$  &  23.00  &  3.90  &  33.07  &  31.93  &  23.84  &  6.19  &  31.38  &  19.15  &  35.34  &  8.24  &  45.43  &  35.28  &  27.39  &  28.79\\
\code{bge-base-en-v1.5}  &  23.44  &  3.86  &  32.79  &  31.47  &  24.17  &  6.37  &  31.80  &  18.68  &  36.59  &  8.47  &  47.44  &  36.87  &  28.07  &  29.01\\
\code{bge-large-en-v1.5}  &  23.12  &  3.76  &  31.93  &  30.67  &  \high{27.14}  &  \high{7.24}  &  \high{35.77}  &  \high{21.16}  &  35.51  &  7.82  &  45.71  &  36.39  &  28.59  &  29.40 \\
\midrule
\code{gte-Qwen2-1.5B-inst.}$^\spadesuit$  &  28.99  &  4.57  &  39.58  &  38.16  &  31.85  &  8.37  &  39.95  &  22.69  &  45.47  &  10.02  &  55.68  &  43.85  &  35.44  &  34.90\\
\code{e5-mistral-7b}  &  25.38  &  4.13  &  34.81  &  33.47  &  28.22  &  7.60  &  35.99  &  22.28  &  42.31  &  9.03  &  51.34  &  40.24  &  31.97  &  32.00\\
\code{GritLM-7B}$^\spadesuit$  &  29.67  &  4.82  &  41.30  &  39.77  &  30.86  &  8.18  &  39.99  &  \high{25.25}  &  48.92  &  10.31  &  59.01  &  46.55  &  36.48  &  37.19 \\
\code{NV-Embed-v1}$^\spadesuit$  &  \high{35.50}  &  \high{5.54}  &  \high{48.36}  &  \high{46.72}  &  \high{33.08}  &  \high{8.77}  &  \high{41.08}  &  24.83  &  \high{51.28}  &  \high{11.13}  &  \high{61.49}  &  \high{48.05}  &  \high{39.95}  &  \high{39.87} \\
\hline
\rowcolor{Gainsboro}\multicolumn{15}{l}{\textit{Cross-encoder re-ranking models}} \\
\code{mxbai-rerank-large-v1}  &  31.75  &  4.79  &  43.82  &  42.20  &  24.84  &  5.96  &  32.13  &  19.53  &  35.65  &  7.31  &  43.38  &  33.43  &  30.75  &  31.72\\
\code{monot5-base-msmarco}  &  26.91  &  4.32  &  37.16  &  35.03  &  30.43  &  8.11  &  38.04  &  20.45  &  46.32  &  9.15  &  56.53  &  46.36  &  34.56  &  33.95\\
\code{bge-reranker-v2-m3}  &  31.25  &  4.92  &  42.97  &  41.34  &  34.04  &  8.86  &  42.85  &  26.46  &  43.81  &  10.38  &  53.28  &  41.84  &  36.36  &  36.54\\
\code{jina-reranker-v2-base}  &  33.31  &  5.06  &  44.20  &  42.93  &  36.79  &  9.49  &  46.20  &  28.55  &  \high{53.18}  &  \high{11.56}  &  \high{63.22}  &  \high{50.70}  &  41.09  &  40.72\\
\code{bge-reranker-v2-gemma}  &  \high{38.59}  &  \high{5.67}  &  \high{50.14}  &  \high{48.67}  &  \high{38.08}  &  \high{10.05} &  \high{47.65}  &  \high{28.98}  &  50.39  &  11.54  &  61.69  &  49.48  &  \high{42.36}  &  \high{42.38} \\
\bottomrule
\end{tabular}
}
\caption{Experiments are conducted under the \textbf{\textit{w/ inst.}} setting, with retrieval performed within each subset individually.}\label{table:category-wo-inst}
\end{table*}

%% file: table/train.tex
\begin{table*}[ht]
\centering
\begin{adjustbox}{width=\columnwidth,center}
\setlength\tabcolsep{4pt}
\begin{tabular}{l cccc cccc cccc |cc}

\toprule
\multirow{2}{*}{\textbf{Model}} & 
\multicolumn{4}{c}{\textbf{\ours-Web}} & 
\multicolumn{4}{c}{\textbf{\ours-Code}} & 
\multicolumn{4}{c}{\textbf{\ours-Customized}} & 
\multicolumn{2}{c}{\textbf{Avg.}} \\
\cmidrule(lr){2-5} \cmidrule(lr){6-9} \cmidrule(lr){10-13}  \cmidrule(lr){14-15}
& N@10 & P@10 & R@10 & C@10
& N@10 & P@10 & R@10 & C@10
& N@10 & P@10 & R@10 & C@10
& N@10 & C@10
\\
\midrule
\code{bge-large-en-v1.5}$^\ddagger$  &  25.07  &  3.94  &  33.52  &  31.85  &  32.05  &  7.98  &  41.03  &  26.11  &  41.72  &  8.80  &  49.26  &  37.99  &  32.95  &  31.98 \\
\code{bge-large-en-v1.5}$^\ddagger$  &  23.41 &  3.22  &  32.02  &  31.97  &  30.74  &  6.31  &  36.03  &  24.31  &  37.31  &  7.30  &  46.31  &  34.21  &  30.49  &  30.16  \\
\code{bge-large-en-v1.5}  &  18.90  &  3.13  &  25.80  &  24.50  &  24.49  &  6.67  &  32.95  &  19.30  &  25.72  &  5.54  &  32.18  &  24.79  &  23.03  &  22.86\\
\midrule
\code{bge-base-en-v1.5}$^\ddagger$  &  20.35  &  3.41  &  28.19  &  26.77  &  30.69  &  7.70  &  39.16  &  24.95  &  37.01  &  8.53  &  47.21  &  36.40  &  29.35  &  29.37\\
\code{bge-base-en-v1.5}$^\dagger$ & 
19.05  &  2.98  &  25.13  &  23.70  &  24.90  &  6.41  &  35.44  &  22.50  &  33.51  &  6.40  &  44.20  &  35.71  & 25.82& 27.30 \\
\code{bge-base-en-v1.5}  &  17.76  &  2.91  &  23.59  &  22.20  &  22.44  &  6.03  &  29.96  &  17.29  &  25.98  &  5.71  &  32.17  &  24.26  &  22.06  &  21.25\\
\midrule
\code{e5-large-v2}$^\ddagger$  &  23.15  &  3.74  &  31.41  &  29.94  &  33.05  &  7.79  &  40.42  &  26.97  &  34.33  &  6.60  &  42.08  &  34.26  &  30.18  &  30.39\\
\code{e5-large-v2}$^\dagger$  &  21.33  &  2.94  &  28.34  &  25.13  &  30.45  &  6.45  &  38.20  &  26.70  &  30.13  &  5.60  &  39.82  &  32.33  &  27.30  & 28.05 \\
\code{e5-large-v2}  &  17.03  &  2.67  &  21.77  &  20.63  &  18.94  &  4.90  &  25.95  &  16.26  &  26.37  &  6.07  &  32.19  &  23.17  &  20.78  &  20.02\\
\midrule
\code{e5-base-v2}$^\ddagger$  &  19.97  &  3.33  &  27.67  &  26.36  &  26.45  &  5.92  &  32.71  &  22.24  &  31.03  &  6.06  &  38.42  &  30.92  &  25.81  &  26.51\\
\code{e5-base-v2}$^\dagger$ &  15.44  &  2.78  &  25.37  &  23.61  &  24.50  &  5.20  &  30.12  &  19.37  &  28.03  &  5.47  &  40.21  &  31.92  &   22.66  &  24.97 \\
\code{e5-base-v2}  &  14.42  &  2.46  &  19.18  &  18.00  &  19.80  &  5.04  &  25.89  &  15.37  &  22.69  &  5.11  &  29.13  &  22.25  &  18.97  &  18.54\\
\bottomrule
\end{tabular}
\end{adjustbox}
\caption{
Experimental results of IR models before and after training on our datasets.
Models trained with the concatenation of instruction and query are denoted by $\ddagger$.
In contrast, the variants trained solely on the query as input are marked with $\dagger$ (See the ablation study in \S~\ref{sec:train} for details).}
\label{table:train}
\end{table*}

%% file: table/pass-rate.tex
\begin{table}[htbp]
\centering
\begin{adjustbox}{width=\columnwidth,center}
\begin{tabular}{@{} l l ll ll ll@{}}
\toprule
\multirow{2}{*}{\textbf{Dataset}} & 
\multicolumn{1}{c}{\textbf{\ours}} & 
\multicolumn{2}{c}{\textbf{ToolBench-G1}} & 
\multicolumn{2}{c}{\textbf{ToolBench-G2}} & 
\multicolumn{2}{c}{\textbf{ToolBench-G3}} \\
\cmidrule(lr){2-2} \cmidrule(lr){3-4} \cmidrule(lr){5-6}  \cmidrule(lr){7-8} 
&  NDCG@10 
& NDCG@10  &  Pass Rate 
& NDCG@10 & Pass Rate 
& NDCG@10 & Pass Rate \\
\hline

\rowcolor{Gainsboro}\multicolumn{8}{l}{\textit{gpt-3.5-turbo as tool-use LLM}} \\
oracle 
&  -  &  - &  62.00 &   - &  57.20  &  - &  67.40 \\
\midrule
bge-large-en-v1.5 
&  23.03  & 34.29 &  50.60  & 9.48 &  49.00  & 29.69  &  56.90 \\
bge-large-en-v1.5 $^\spadesuit$ 
&  32.95$_{\uparrow 43.07\%}$ & 71.11 $_{\uparrow 107.38\%}$ &  59.50$_{\uparrow 17.59\%}$ & 18.11$_{\uparrow 91.03\%}$ &  58.40$_{\uparrow 19.18\%}$   & 67.87$_{\uparrow 128.60\%}$ &  59.20$_{\uparrow 4.04\%}$  \\
\midrule
bge-base-en-v1.5
&  22.06 & 36.89 &  50.60  & 9.28 & 51.20  & 33.02 &   57.70 \\
bge-base-en-v1.5$^\spadesuit$
& 29.35$_{\uparrow 33.05\%}$ & 67.52$_{\uparrow 83.03\%}$ &  56.60$_{\uparrow 11.86\%}$ & 16.01$_{\uparrow 72.52\%}$ &  59.60$_{\uparrow 16.41\%}$   & 60.75$_{\uparrow 83.98\%}$ &  60.80$_{\uparrow 5.37\%}$  \\
\midrule
e5-large-v2
&  20.78 &  44.91 &  47.50  &11.57 &  56.50  & 43.43 &  55.70 \\
e5-large-v2$^\spadesuit$
&  30.18$_{\uparrow 45.24\%}$ & 70.08$_{\uparrow 56.05\%}$ &  57.00$_{\uparrow 20.0\%}$  & 17.71$_{\uparrow 53.07\%}$ &  62.10$_{\uparrow 9.91\%}$   & 66.09$_{\uparrow 52.18\%}$ &  58.00$_{\uparrow 3.99\%}$  \\
\midrule
e5-base-v2
&  18.97 & 38.66 &  49.60  & 9.87 &   54.10  & 37.35  &  54.20 \\
e5-base-v2$^\spadesuit$
&  25.81$_{\uparrow 36.06\%}$ &65.79$_{\uparrow 70.18\%}$  &  56.90$_{\uparrow 14.72\%}$ & 17.45$_{\uparrow 76.80\%}$  &  60.80$_{\uparrow 12.38\%}$   &62.74$_{\uparrow 67.98\%}$  &  62.40$_{\uparrow 15.13\%}$  \\

\hline
\rowcolor{Gainsboro}\multicolumn{8}{l}{\textit{ToolLlama as tool-use LLM}} \\
oracle 
&  -  &  - &  53.6 	 &   - & 50.8  &  - &  49.1\\
\midrule
bge-large-en-v1.5  &  23.03  & 34.29 & 37.60   & 9.48 &  41.30  & 29.69  & 37.20   \\
bge-large-en-v1.5$^\spadesuit$ &  32.95$_{\uparrow 43.07\%}$ & 71.11 $_{\uparrow 107.38\%}$ &  45.10$_{\uparrow 19.95\%}$ & 18.11$_{\uparrow 91.03\%}$ &   47.30$_{\uparrow 14.53\%}$  & 67.87$_{\uparrow 128.60\%}$ & 39.60$_{\uparrow 6.45\%}$  \\
\midrule
bge-base-en-v1.5 &  22.06 & 36.89 &  47.80  & 9.28 &   46.10 & 33.02 & 36.10  \\
bge-base-en-v1.5$^\spadesuit$ & 29.35$_{\uparrow 33.05\%}$ &   67.52$_{\uparrow 83.03\%}$  & 50.60$_{\uparrow 5.86\%}$  & 16.01$_{\uparrow 72.52\%}$ &  49.80$_{\uparrow 8.03\%}$ & 60.75$_{\uparrow 83.98\%}$ &  45.70$_{\uparrow 26.60\%}$ \\
\midrule
e5-large-v2 &  20.78 &  44.91 & 41.50 & 11.57 & 46.60  & 43.43 & 40.20     \\
e5-large-v2$^\spadesuit$ &  30.18$_{\uparrow 45.24\%}$ & 70.08$_{\uparrow 56.05\%}$ &   44.50$_{\uparrow 7.23\%}$  & 17.71$_{\uparrow 53.07\%}$ &   49.80$_{\uparrow 4.72\%}$ & 66.09$_{\uparrow 52.18\%}$ & 43.80$_{\uparrow 4.50\%}$ \\
\midrule
e5-base-v2 &  18.97 & 38.66 &  42.20 & 9.87 & 45.20  & 37.35  & 42.00 \\
e5-base-v2$^\spadesuit$ &  25.81$_{\uparrow 36.06\%}$ &65.79$_{\uparrow 70.18\%}$  & 49.10$_{\uparrow 16.35\%}$  & 17.45$_{\uparrow 76.80\%}$  &  48.30$_{\uparrow 6.86\%}$ & 62.74$_{\uparrow 67.98\%}$  & 44.70$_{\uparrow 6.60\%}$  \\
\bottomrule
\end{tabular}
\end{adjustbox}
\caption{Experiment results of IR models before and after training. We also show the end-to-end task pass rate of tool-use LLMs when equipped with the tools retrieved by the IR models on ToolBench dataset.}\label{tab:pass-rate}
\end{table}